\pdfoutput=1

\PassOptionsToPackage{table}{xcolor}
\documentclass[11pt]{article}

\usepackage[preprint]{acl}

\usepackage{times}
\usepackage{latexsym}
\usepackage{hyperref}
\usepackage{url}
\usepackage{booktabs}
\usepackage{subcaption}
\usepackage{tabularx}
\usepackage{graphicx}
\usepackage{dsfont}
\usepackage[misc]{ifsym}

\usepackage[most]{tcolorbox}

\usepackage{tcolorbox}

\makeatletter
\newcommand{\DrawLine}{%
  \begin{tikzpicture}
  \path[use as bounding box] (0,0) -- (\linewidth,0);
  \draw[color=blue!75!black,dashed,dash phase=2pt]
        (0-\kvtcb@leftlower-\kvtcb@boxsep,0)--
        (\linewidth+\kvtcb@rightlower+\kvtcb@boxsep,0);
  \end{tikzpicture}%
  }
\makeatother

\usepackage{multirow}
\usepackage[normalem]{ulem}
\useunder{\uline}{\ul}{}

\usepackage[T1]{fontenc}

\usepackage[utf8]{inputenc}

\usepackage{microtype}

\usepackage{inconsolata}

\usepackage{graphicx}

\title{Critic-CoT: Boosting the Reasoning Abilities of Large Language Model via Chain-of-Thought Critic}

\author{
\begin{tabular}{@{}l}
Xin Zheng${}^{1,2}$\thanks{~ This work was done when Xin Zheng interned at Xiaohongshu.} \quad Jie Lou${}^{3}$\textsuperscript{\Letter} \quad Boxi Cao ${}^{1,2}$ \quad Xueru Wen ${}^{1,2}$ \quad Yuqiu Ji${}^{3}$ \\
Hongyu Lin${}^{1}$\textsuperscript{\Letter} \quad Yaojie Lu${}^{1}$ \quad Xianpei Han${}^{1}$ \quad Debing Zhang${}^{3}$ \quad Le Sun${}^{1}$  \\
\end{tabular}\\
${}^{1}$ Chinese Information Processing Laboratory, Institute of Software, \\ Chinese Academy of Sciences
${}^{2}$ University of Chinese Academy of Sciences \\
${}^{3}$ Xiaohongshu Inc \\
{\tt \{zhengxin2020,boxi2020,wenxueru2022\}@iscas.ac.cn} \\
{\tt \{hongyu,luyaojie,xianpei,sunle\}@iscas.ac.cn} \\
{\tt \{yinyue2,dengyang\}@xiaohongshu.com } \\
\footnotetext[2]{Corresponding authors.}
}

\begin{document}
\maketitle
\begin{abstract}
Self-critic has become a crucial mechanism for enhancing the reasoning performance of LLMs. 
However, current approaches mainly involve basic prompts for intuitive instance-level feedback, which resembles System-1 processes and limits the reasoning capabilities. Moreover, there is a lack of in-depth investigations into the relationship between LLM's ability to criticize and its task-solving performance. 
To address these issues, we propose Critic-CoT, a novel framework that pushes LLMs toward System-2-like critic capability.
Through a step-wise CoT reasoning paradigm and the automatic construction of weak-supervision data without human annotation, Critic-CoT enables LLMs to engage in slow, analytic self-critique and refinement, thereby improving their reasoning abilities.
Experiments on GSM8K and MATH and out-of-domain evaluation demonstrate that our enhanced model significantly boosts task-solving performance by filtering out invalid solutions or iterative refinement. Furthermore, we investigate the intrinsic correlation between critique and task-solving abilities within LLMs, discovering that these abilities can mutually reinforce each other rather than conflict.
\footnote{Our code and data are available at \url{https://github.com/AlignRM/Critic-CoT}}
\end{abstract}

\section{Introduction}

Enhancing the reasoning abilities of large language models is essential for creating more intelligent and reliable AI systems, which has drawn extensive attention from researchers~\citep{chollet2019measureintelligence, bubeck2023sparksartificialgeneralintelligence, morris2024position}. From a cognitive perspective, the procedure of human reasoning involves constant reflection and revision~\citep{hegel1991encyclopaedia, d0f1f802-4f88-3f86-ab42-b3525806553e, Popper1934-POPTLO-2}, which has inspired increasing focus on integrating self-critic mechanisms in the reasoning process of large-scale models~\citep{kim2023language, shinn2023reflexion, madaan2023selfrefine}. This involves iteratively allowing the model to generate feedback on its own responses and then refining its reasoning based on the feedback. 
Compared with traditional critic methods that depend on feedback from external sources~\citep{saunders2022selfcritiquingmodelsassistinghuman, mcaleese2024llmcriticshelpcatch}, self-critic relies solely on the model's internal capabilities, thus reducing the high cost of additional human annotation, and serving as a promising potential solution to scalable oversight~\citep{leike2018scalableagentalignmentreward, burns2023weaktostronggeneralizationelicitingstrong, cao2024scalableautomatedalignmentllms}.

\begin{figure*}[t]
    \centering
    \includegraphics[width=\textwidth]{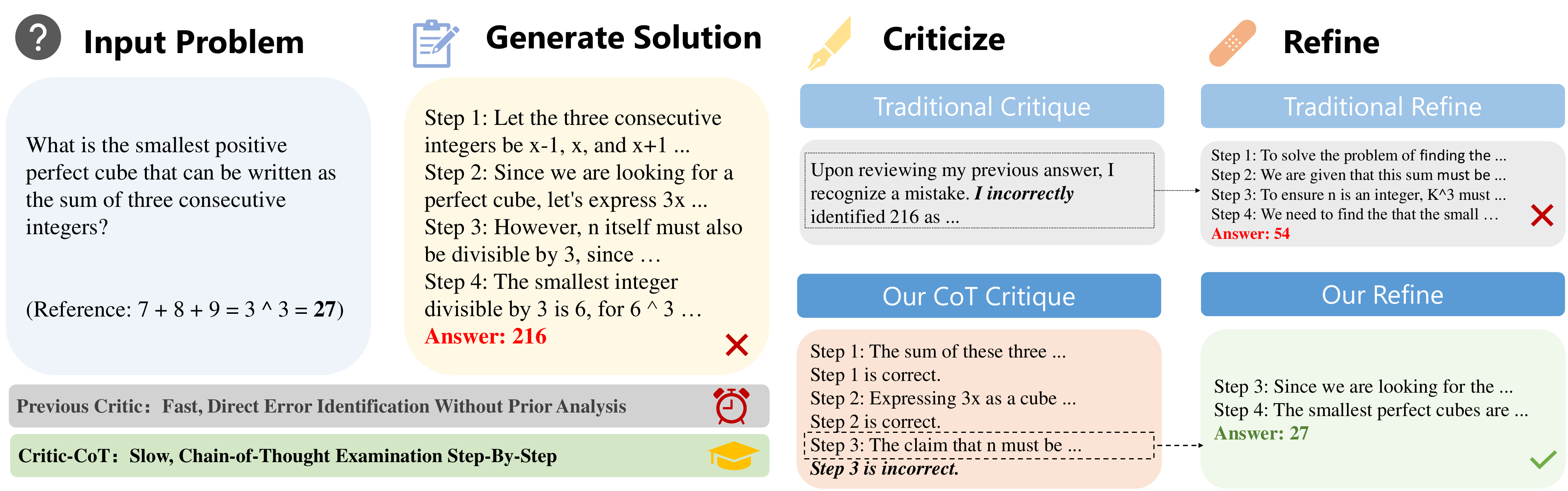}
    \caption{Illustration of Critic-CoT: Previous instance-level critic methods attempt to identify errors directly without any prior analysis, and restart from the beginning during refinement. In contrast, our proposed Critic-CoT framework performs a step-wise examination using the Chain-of-Thought approach. When refining, rather than starting from scratch, our method makes the correction from the specific error step with the help of the corresponding critique.}
    \label{fig:critic-cot-figure1}
\end{figure*}

However, current studies primarily focus on utilizing LLMs' critique abilities to enhance their performance. Yet, relatively little attention has been given to the investigation and development of the critique ability itself.
Firstly, existing critique methods are often overly simplistic, typically relying on a basic prompt to directly point out the error, without stepwise Chain-of-Thought examination or training procedure, which leads to relatively poor self-critic accuracy \citep{luo2023critiqueabilitylargelanguage, west2024the}. 
Specifically, proposing a valid critique is a complicated task that requires a thorough understanding of statements and precise negativity. However, current LLMs are normally not explicitly trained for critic capability. 
Therefore, these simple approaches usually tend to ``criticize'' like System-1 (fast thinking mode), which is more intuitive and likely to make mistakes, rather than more rigorous and deliberate System-2 (slow thinking mode) \citep{kahneman2011thinking, yu2024distilling21}, while shifting LLMs from System-1 toward System-2, which is to perform systematical analysis that fully utilizes the advantage of Chain-of-Thought reasoning, emerges as a promising approach for improving the reasoning capability~\citep{openaio1}.
This limitation diminishes the effectiveness of self-critic and, further, self-correct \citep{huang2024large}.
Secondly, the capabilities of task-solving and self-critic are both dependent on the model's inherent knowledge, while there is currently a lack of in-depth exploration regarding the correlation between these two capabilities within LLMs. In that case, it's challenging to balance the task-solving and the self-critic capabilities of the model within the self-critic framework, which poses a significant obstacle to the subsequent development. 

To this end, this paper is devoted to diving into the following critical research questions:
\begin{itemize}
    \item How can we enhance a model’s critique ability, pushing it toward System 2 reasoning?
    \item What is the relationship between a model’s critique ability and its task-solving capability?
\end{itemize}
To answer the above questions, as shown in Figure \ref{fig:critic-cot-figure1}, 
we propose Critic-CoT, a novel framework designed to enhance LLMs' reasoning abilities. Through step-wise Chain-of-Thought critique format and automated data construction through weak supervision, our method is able to strengthen System-2-like critic ability, without the intensive cost of human annotation.
Specifically, during training, we let LLMs criticize and refine their solutions in a complete CoT way, and collect successful pairs that convert wrong solutions into correct ones, or affirm the validity of original right solutions. 
After supervised fine-tuning on the obtained step-wise critic-refine data, we enable the target LLM to analyze and criticize each step of its generated reasoning procedure, so that it can filter out wrong attempts and preserve the correct ones with greater precision. 
During inference, to leverage the model's abilities of CoT-critique and refinement, we employ two strategies: 1) majority vote filtering involves using the critic model to evaluate multiple generated solutions and filter out those incorrect; 
and 2) iterative refinement, on the other hand, involves repeatedly critiquing and refining a solution until no further error is detected.

Through a series of experiments on the in-domain dataset of GSM8K and MATH, together with out-of-domain evaluation on StrategyQA, AGIEval and HumanEval, we find that our trained critic model can fairly distinguish incorrect solutions from correct ones, and improve the reasoning accuracy via iterative refinement or critic filtering. These results demonstrate the helpfulness and effectiveness of our proposed method. 
Additionally, we observed that our critic model already exhibits noticeable performance improvements in task-solving,
even in the absence of additional critique steps during the decoding phase. Such findings reveal that strengthening the ability to critique and refinement would not compromise the task-solving performance, but improve it. 
This also suggests the presence of an intrinsic mechanism by which critique ability and task-solving capability mutually reinforce one another.

Our main contributions are as follows:
\begin{itemize}
    \item We propose Critic-CoT, which pushes the critic paradigm of LLMs from System-1-like incentive ``thinking'' toward System-2-like deliberate ``reasoning''. 
    \item Through experiments, we find that Critic-CoT can effectively teach the model to criticize and refine its own output step by step, thus noticeably improving the reasoning performance.
    \item Moreover, we find that for LLMs, the ability of critique and refinement could mutually reinforce, which may shed light on designing more advanced self-critic framework designs in future work.

\end{itemize}

\section{Related Works}





With the development of LLMs, self-critic has emerged as a widely-adopted mechanism for reasoning, code generation, computer control, hallucination mitigation, retrieval-augmented generation and other tasks \citep{kim2023language, shinn2023reflexion, madaan2023selfrefine, ji-etal-2023-towards, asai2024selfrag}. However, typical self-critic approaches like Reflexion \citep{shinn2023reflexion}, only utilize the LLMs' existing critique ability without further enhancement, and require the gold environment signal to iteratively generate critiques and make refinements. As long as external feedback is not available, off-the-shelf LLMs cannot perform intrinsic self-correct effectively due to limited critique and refinement abilities \citep{huang2024large, luo2023critiqueabilitylargelanguage, DBLP:journals/corr/abs-2312-17080}. 
Later, several works are proposed to improve self-reflection via a carefully designed prompting pipeline on frozen LLMs, and no active training process is involved \citep{zhang-etal-2024-self-contrast,yan-etal-2024-mirror,wu-etal-2024-large}. Concurrently, \citet{zhang2024generativeverifiersrewardmodeling} trained a generative reward model on the outcome level rather than the process level, and did not incorporate refinement into the schema. Therefore, given the limited critic ability of current LLMs, how to train a robust and applicable critic model, which conducts detailed Chain-of-Thougt analysis in a step-wise systematic manner, and thus shifts from System-1 reasoning toward more deliberate System-2 reasoning \citep{kahneman2011thinking}, is worth investigating. 

From the perspective of recursive reward modeling \citep{leike2018scalableagentalignmentreward, saunders2022selfcritiquingmodelsassistinghuman} and scalable oversight \citep{burns2023weaktostronggeneralizationelicitingstrong}, \citet{mcaleese2024llmcriticshelpcatch} recently trained ``CriticGPT'' to assist human labelers, which aims to improve the ability of human rather than the base model, i.e., improve the overall recall of error detection, rather than precision. While in this paper, we try to improve the reasoning ability of LLM without costly human annotation.

\section{Method}
\begin{figure*}[t]
    \centering
    \includegraphics[width=\textwidth]{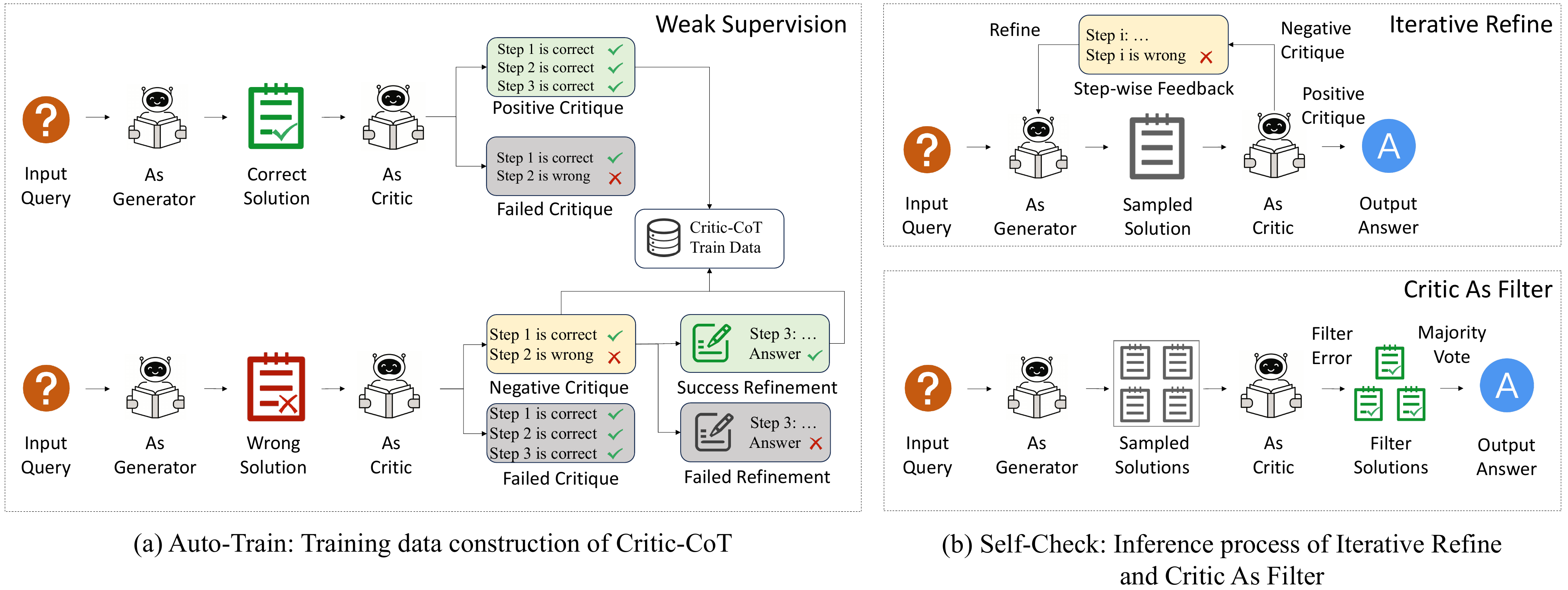}
    \caption{The Process of Critic-CoT during training (a) and inference (b). For training, we collect the critic-refine data on the generator's samples via weak supervision (Section \ref{section:Chain-of-Thought Critique}). Through fine-tuning, we enable the target model to criticize and refine its own reasoning process. Then, during inference, we can leverage the capabilities via Iterative Refine or Critic As Filter (Section \ref{section:Inference}). }
    \label{fig:critic_process}
\end{figure*}

Critic CoT is to equip LLMs with the ability to criticize and refine themselves step-by-step. As shown in Figure \ref{fig:critic_process}, it consists of two modules, including automated data construction via weak-supervision and self-check at inference-time. The weak-supervision principles are in Section \ref{section:Chain-of-Thought Critique}, followed by the training process in Section \ref{section:Chain-of-Thought Critique}, and the inference strategies in \ref{section:Inference}.

\subsection{Chain-of-Thought Critique}
\label{section:Chain-of-Thought Critique}

In this work, we utilize a step-wise chain-of-thought critique, which makes the critique-refine process both controllable and formalizable, thereby facilitating the collection of weak supervision data.
Formally, given the question $Q$ and the corresponding gold answer $Ans$, we have the $n$-step attempt $Att = [s_1, ..., s_n]$ with predicted answer $Pred$ sampled by generator $G$.
The corresponding critique $Cri$ then can be represented as $L = [l_1, ..., l_n]$, where the step label $l_i = +1$ indicates that step $i$ is predicted to be correct, and $l_i = -1$ to be incorrect.
Then the refinement $Att' = [s'_i, ..., s'_{n'}]$ is start from the first incorrect step $i$ with new answer $Pred'$.
We automatically annotate the process labels as follows:
\begin{itemize}
    \item $Pred \neq Ans, -1 \notin L$:
    The attempt is wrong, yet the critique did not discover any error step. Thus the critique itself is problematic, and we need to sample another critique.
    \item $Pred \neq Ans, -1 \in L, Pred' \neq Ans$:
    The attempt is wrong, and the critique found an error, but still, the refinement is not correct. There could be two cases for this situation: (1) the refinement is unsuccessful; (2) the critique did not detect an earlier mistake. We simply sample another critique and corresponding refinement for this situation.
    \item $Pred \neq Ans, -1 \in L, Pred' = Ans$:
    Not only did the critique point out the error, but also the refinement reached the correct answer. We then believe the critique is valid, and collect the critique data instance $C=(Q, Att, Cri)$ and the refinement data $R=(Q, Att, Cri_{-1}, Att')$, where $Cri_{-1}$ is the critique of last step, since explaining why previous steps are correct may not be helpful.
    \item $Pred = Ans, -1 \notin L$:
    The attempt is correct, and the critique believes it is correct. So we can collect the positive critique data instance $C=(Q, Att, Cri)$.
    \item $Pred = Ans, -1 \in L$:
    The attempt reached the correct answer, yet the critique found an error. Then, the critique could be wrong, and we need to sample another critique.
\end{itemize}

\subsection{Auto Train: Two-Stage Training}
\label{section:Two-Stage Training}
To enable the model to acquire self-critiquing and refining capabilities, we first need to provide it with basic critiquing abilities, followed by self-critique for further enhancement. 
The overall training procedure is divided into two stages.

\paragraph{Stage 1} In the first step, we collect high-quality critique data to provide the model's basic critiquing ability. 
Specifically, we first sample both positive and negative solutions from a representative instruction-following model $\mathcal{M}_G$ on the dataset $D$. 
Then, we utilize LLMs like GPT4-Turbo to serve as critic model $M_C$.
For each generated attempt $Att$, the critic model will retry at most $k$ times to produce a valid critique until it reaches one of the weak supervision constraints. 
This will form the critic-refine dataset $D_1 = \{(Q,Att,Cri)\} \bigcup \{(Q,Att,Cri_{-1},Att^\prime)\}$ for fine-tuning the initial model $\mathcal{M}_0$ into the critic model $\mathcal{M}_1$.
Note that in this process, we actually distill \texttt{Pass1@N} of the teacher model $\mathcal{M}_C$ into \texttt{Top1@N} of the student model.
So, the theoretical upper bound of the student model is not necessarily limited by the teacher model’s performance.

\paragraph{Stage 2} In the second step, we leverage the model's self-critique to enhance its critiquing and refining capabilities further.  
Namely, we let the learned critic model $\mathcal{M}_1$ criticize and refine its own output. 
We first sample $M$ correct-answer solutions and $M$ incorrect-answer solutions for each question $Q$ in the original dataset $D$.
Then, for each attempt $Att$, we employ $\mathcal{M}_1$ to repeatedly criticize and refine at most $k$ times. 
If the model fails to critique even after $k$ times, we fall back on the critique from a stronger yet frozen model $M_C$ as the final choice.
Finally, we collect dataset $D_2=\{(Q,Att,Cri)\} \bigcup \{(Q,Att,Cri_{-1},Att^\prime)\}$ and use $D_1 \bigcup D_2$ to train the initial model $\mathcal{M}_0$ into the final critic model $\mathcal{M}_2$, which is similar to \citet{wang2024selftaughtevaluators}.
This procedure helps the model to learn to criticize and refine its own reasoning outputs better.

\begin{table*}[ht]
\centering
\small
\begin{tabularx}{\textwidth}{llll}
\toprule
\textbf{Model}   & \textbf{Sampling Method} & \textbf{GSM8K}   & \textbf{MATH500} \\
\midrule
\textit{Base Model}\\
Llama-3-70B-Instruct \citep{llama3modelcard} & - & 89.6 & 50.4 \\
 & Maj1@96 & 94.1 & 62.2 \\
 & Maj1@512 & - & 63.4 \\
Llama-3.1-70B-Instruct \citep{llama3modelcard} & - & 94.5 & 65.7 \\
GPT4-0314 \citep{DBLP:journals/corr/abs-2303-08774} & - & 92.0 & 52.6 \\
DeepSeek-V2 Chat-236B \citep{DBLP:journals/corr/abs-2405-04434} & - & 92.2 & 56.3 \\
\midrule
\textit{Reasoning-Enhanced Model}\\
MATH-Minos, Mistral-7B \citep{DBLP:journals/corr/abs-2406-14024} & PRM+Maj1@256 & 87.8 & 38.6 \\
InternLM-MATH-20B \citep{DBLP:journals/corr/abs-2402-06332} & PRM Best-of-100 & 89.3 & 50.0 \\
DART, Llama3-70B \citep{tong2024dartmathdifficultyawarerejectiontuning} & - & 89.6 & 56.1 \\
Math-Shepherd, DeepSeek-67B \citep{DBLP:journals/corr/abs-2312-08935} & PRM+Maj1@256 & 92.5 & 48.1 \\
\midrule
\textit{Ours}\\
\textbf{Critic-CoT, Llama-3-70B-Instruct} & - & 91.7 & 57.6  \\
 & Iterative Refine & \textbf{93.3} {\color[HTML]{32cb00}$\uparrow$ 1.6} & \textbf{57.8} {\color[HTML]{32cb00}$\uparrow$ 0.2} \\
 & Maj1@96 & 94.8 & 64.6 \\
 & Critic + Maj1@96 & \textbf{95.4} {\color[HTML]{32cb00}$\uparrow$ 0.6} & 66.6 {\color[HTML]{32cb00}$\uparrow$ 2.0} \\
 & Maj1@512 & - & 65.4 \\
 & Critic + Maj1@512 & - & \textbf{68.4} {\color[HTML]{32cb00}$\uparrow$ 3.0} \\
\bottomrule
\end{tabularx}
\caption{Solution Accuracy of GSM8K and MATH500. Compared with the base model, iterative refinement with our trained model improves from 89.6\% to 93.3\%\% for GSM8K and from 50.4\% to 57.8\% for MATH500, while the critic filter increases the accuracy to 95.4\% for GSM8K and 68.4\% for MATH500.}
\label{table:Solution Accuracy of GSM8K and MATH500}
\end{table*}

\subsection{Inference: Self-Check}
\label{section:Inference}

\paragraph{Iterative Refine}
\label{sec:Iterative Refine}
One single-turn refinement, which consists of multiple steps, may still contain errors. Therefore, we could iteratively inspect the refined solution, and re-refine once the critique found a mistake, and only output the final solution if it's convincing for the critic, or if it reached the maximum retry. To avoid de-generation after too many refinements, we set the maximum refine depth $d=8$, and restart from the initial solution after $d$ unsuccessful refinement at most $n=8$ times. Figure \ref{fig:example_MATH} presents a single successful round of critique and refinement.

\paragraph{Critic As Filter} 
\label{sec:Critic As Filter}
Self-consistency is an effective way to reduce variance. With the ability to critique, we can filter out predict-to-be-wrong answers to further boost the performance. Specifically, for the $m$ attempts $S=\{(Att, Pred)\}$, we first let our model $\mathcal{M}$ check each attempt and obtain the stepwise label, which is $S_c=\{(Att, Pred, L)\}$.
Then those which detect the error at some step are filtered out and reach $S^\prime_c = \{(Att, Pred, L) | -1 \notin L\}$.
Finally, we perform the majority vote to get the answer.

\section{Experiment}
We apply the Critic-CoT training process on the training dataset of GSM8K and MATH (Section \ref{section:setup}), and observe a noticeable performance improvement (Section \ref{section:Main Results}), and out-of-domain evaluations on AGIEval, StrategyQA, and HumanEval further exhibits the generalization of our trained critic ability (Section \ref{section:Out-of-Domain Results}) For more analysis, discussion, see Appendix \ref{appendix:analysis} and Appendix \ref{appendix:discussion}, and the prompt is presented in Appendix \ref{appendix:Prompts}.

\begin{table}[!h]

    \begin{subtable}[h]{0.45\textwidth}
        \centering
        \small
        \begin{tabular}{l l}
        \toprule
        \textbf{Model} & \textbf{Acc.} \\ \midrule
        Llama-3-70B-Instruct & 56.6 \\
        Llama-3.1-70B-Instruct & 61.8 \\
        DeepSeek-V2 Chat-236B & 61.4 \\
        GPT4o & \underline{65.2} \\
        \midrule
        \textbf{Critic-CoT, GSM8K} & 54.7  \\
        - Iterative Refine & 55.6  {\color[HTML]{32cb00}$\uparrow$ 0.8} \\
        - Maj1@96 & 60.7  \\
        - Critic + Maj1@96 & 60.3  {\color[HTML]{FF0000}$\downarrow$ 0.4} \\        
        \textbf{Critic-CoT, MATH} & 59.8 \\
        - Iterative Refine & \textbf{63.7} {\color[HTML]{32cb00}$\uparrow$ 3.9} \\
        - Maj1@96 & 61.0 \\
        - Critic + Maj1@96 & \textbf{61.2} {\color[HTML]{32cb00}$\uparrow$ 0.2} \\ \bottomrule
        \end{tabular}
        \caption{AGIEval}
        \label{subtable:Solution Accuracy of AGIEval}
    \end{subtable}
    \hfill
    \begin{subtable}[h]{0.45\textwidth}
        \centering
        \small
        \begin{tabular}{l l}
        \toprule
        \textbf{Model} & \textbf{Acc.} \\ \midrule
        Llama-3-70B-Instruct & 76.2  \\
        Llama-3.1-70B-Instruct & \underline{84.3}  \\
        DeepSeek-V2 Chat-236B & 75.6   \\
        GPT4-0314 & 83.6  \\
        \midrule
        \textbf{Critic-CoT, GSM8K} & 77.5  \\
        - Iterative Refine & 78.8  {\color[HTML]{32cb00}$\uparrow$ 1.3} \\
        - Maj1@96 & 78.7   \\
        - Critic + Maj1@96 & \textbf{80.5}  {\color[HTML]{32cb00}$\uparrow$ 1.8 } \\
        
        \textbf{Critic-CoT, MATH} & 78.0  \\
        - Iterative Refine & \textbf{80.1}  {\color[HTML]{32cb00}$\uparrow$ 2.1} \\
        - Maj1@96 & 78.3   \\
        - Critic + Maj1@96 & 79.7  {\color[HTML]{32cb00}$\uparrow$ 1.4 } \\ \bottomrule
        \end{tabular}
        \caption{StrategyQA}
        \label{subtable:Solution Accuracy of StrategyQA}
     \end{subtable}
         \hfill
    \begin{subtable}[h]{0.45\textwidth}
        \centering
        \small
        \begin{tabular}{l l}
        \toprule
        \textbf{Model} & \textbf{Pass@1} \\ \midrule
        Llama-3-70B-Instruct & 76.2  \\
        Llama-3.1-70B-Instruct & 80.5  \\
        DeepSeek-V2 Chat-236B & 81.1    \\
        GPT4-0314 & \underline{86.6}   \\
        \midrule
        \textbf{Critic-CoT, GSM8K} & 77.4  \\
        - Iterative Refine & \textbf{78.1}   {\color[HTML]{32cb00}$\uparrow$ 0.7} \\
        
        \textbf{Critic-CoT, MATH} & 84.1   \\
        - Iterative Refine & \textbf{84.8 }  {\color[HTML]{32cb00}$\uparrow$ 0.7} \\
        \bottomrule
        \end{tabular}
        \caption{HumanEval}
        \label{subtable:Solution Accuracy of HumanEval}
     \end{subtable}
     
     \caption{Solution Accuracy of standardize exam dataset AGIEval (\ref{subtable:Solution Accuracy of AGIEval}), multi-hop reasoning dataset StrategyQA (\ref{subtable:Solution Accuracy of StrategyQA}) and code generation dataset HumanEval (\ref{subtable:Solution Accuracy of HumanEval}). Our models generally show robust generalization.}
     \label{tab:OOD solution}
\end{table}

\subsection{Setup}
\label{section:setup}
\subsubsection{Model}
We fine-tune the critic-refine model on Llama-3-70B-Instruct \citep{llama3modelcard}, which was pre-trained on more than 15 Trillion tokens and has a context length of 8,192. For critique / refinement sampling, we use GPT4-Turbo \citep{DBLP:journals/corr/abs-2303-08774} of the version gpt-4-0125-preview. We use the Huggingface Transformers \citep{wolf-etal-2020-transformers}, DeepSpeed \citep{10.1145/3458817.3476205} and FastChat \citep{NEURIPS2023_91f18a12} libraries for training. We use vLLM library \citep{10.1145/3600006.3613165} for model inference, adapting top-p sampling of $p = 0.95$, with temperature 0.7 for solution sampling, which follows \citet{DBLP:journals/corr/abs-2110-14168}, and 0.5 for critique and refinement. All inferences are zero-shot.

\subsubsection{Dataset}
 
\paragraph{Train \& In-Domain Eval} Separately, we train our model on the problem of GSM8K \citep{DBLP:journals/corr/abs-2110-14168} and MATH \citep{hendrycks2021measuring}. GSM8K is a grade-school-level math word problem dataset, with 7,473 training instances and 1,319 test instances. MATH is a challenging high school math competition dataset, which consists of 7,500 training problems and 5,000 test problems. For the MATH dataset, we also follow the data split of \citet{lightman2024lets}, which adds 4,500 test problems into a training set and, therefore, contains 12,000 training instances and 500 representative test instances. More details are in Appendix \ref{appendix:more_details}.

\paragraph{Out-of-Domain Eval} 
To further evaluate our critic model’s generalization capabilities beyond math, we assess its performance on reasoning tasks using the StrategyQA, AGIEval, and HumanEval datasets, which cover different domains.
StrategyQA \citep{geva-etal-2021-aristotle} is a multi-step reasoning task constructed from Wikipedia, with binary answers indicating either true or false.
AGIEval \citep{zhong2023agievalhumancentricbenchmarkevaluating} comprises standardized exam questions from various fields, including college entrance exams, law school admission tests, math competitions, and lawyer qualification tests. Given the overlap with the MATH test set, we evaluated our model using the original 7,500/5,000 split from MATH, rather than the extended 12,000/500 split.
HumanEval \citep{chen2021evaluatinglargelanguagemodels} contains 164 handwritten Python program problems, which evaluate the code generation capability.

\subsection{Critic-CoT Improves Mathematical Reasoning}
\label{section:Main Results}


The results of in-domain evaluation are shown in Table~\ref{table:Solution Accuracy of GSM8K and MATH500}, which demonstrate \textbf{the effectiveness of Critic-CoT in improving the model's mathematical reasoning performance.}
First, strengthening the ability to critique and refinement would not compromise the task-solving performance, but improve it. After Critic-CoT training, our model's top-1 accuracy increases from 89.6\% to 91.7\% on GSM8K, and from 50.4\% to 57.6\% on MATH500. 
Second, the step-wise self-critique ability of the models can further enhance the reasoning performance during inference, via Iterative Refinement and Critic As Filter. With Iterative Refinement, our model achieves 93.3\% accuracy on GSM8K and 57.8\% on MATH500. Applying Critic As Filter, the performance improves further on the basis of majority vote. On GSM8K it rises from 94.8\% with Maj1@96 to 95.4\% and on MATH500 it rises from 65.4\% with Maj1@512 to 68.4\%.
Thus, our model's accuracy surpasses strong baselines of Process Reward Model assisted model MATH-Minos, InternLM-MATH-20B, Math-Shepherd, and rejection sampling finetuning model DART.
Overall, the results indicate the model's enhanced abilities to identify mistakes and recover from them, thereby boosting the reasoning performance.

\begin{table*}[h]
\centering
\small
\begin{tabularx}{\textwidth}{@{}p{3.5cm}p{3.5cm}p{3.5cm}l@{}}
\toprule
\multirow{2}{*}{\textbf{Data}} & \textbf{Critique of} & \textbf{Refinement of} & \textbf{Critique of} \\
 & \textbf{Wrong Attempt} & \textbf{Wrong Attempt} & \textbf{Correct Answer Attempt} \\ 
\midrule
GSM8K & 86\% & 97\% & 100\% \\
MATH & 85\% & 96\% & 92\% \\ \bottomrule
\end{tabularx}
\caption{Human Evaluation on the critique and refinement of Critic-CoT Training Data. The automatically constructed data maintain high quality, which can well support the critique training process.}
\label{table:Human Evaluation on the Critic-CoT Data}

\end{table*}

\begin{table*}[ht]
\small
\centering
\begin{subtable}[t]{\textwidth}

\begin{tabularx}{\textwidth}{@{}p{3.5cm}lllllllll@{}}
\toprule
\multirow{2}{*}{\textbf{Model}} & \multicolumn{4}{c}{\textbf{Critic}} & \multicolumn{2}{c}{\textbf{Refine}} & \multicolumn{3}{c}{\textbf{Majority Vote}} \\ \cmidrule(lr){2-5} \cmidrule(lr){6-7}  \cmidrule(lr){8-10} 
 & \textbf{P} & \textbf{R} & \textbf{F1} & \textbf{Acc.} & \textbf{Init. Acc} & \textbf{Ref. Acc.} & \textbf{Pass1@N} & \textbf{Maj1@N} & \textbf{+Critic} \\ \midrule
Outcome Label & \textbf{95.5 } & 28.9  & 44.4  & 88.0  & 87.7 & 89.7 & 99.0 & 93.6 & 93.7 \\
Process Label & 67.9 & 22.8 & 34.1 & 89.5 & 88.0 & 89.2 & 99.0 & 93.0 & 93.0 \\ \midrule
Only Refine & 30.0 & 11.4 & 16.6 & 90.8 & \textbf{92.0} & 88.2 & 98.9 & \textbf{95.2} & 95.2 \\
Only Critic & 57.1 & 31.0 & 40.2 & 91.9 & 91.2 & 91.4 & 98.9 & 94.4 & 94.5 \\ \midrule
Stage 1 & 42.5 & 41.5 & 42.0 & 89.3 & 90.7 & 91.1 & 98.9 & 93.6 & 94.2 \\
Stage 2 & 50.0 & 25.0 & 33.3 & 85.5 & 90.5 & 91.3 & 99.0 & 94.4 & 94.4 \\ \midrule
\textbf{Critic-CoT} & 53.3 & \textbf{58.2} & \textbf{55.7} & \textbf{92.3} & 91.7 & \textbf{93.3} & \textbf{99.1} & 94.8 & \textbf{95.4} \\ \bottomrule
\end{tabularx}
\caption{GSM8K}
\label{table:Ablation Study on GSM8K}

\end{subtable}

\begin{subtable}[t]{\textwidth}
\centering
\begin{tabularx}{\textwidth}{@{}p{3.5cm}lllllllll@{}}
\toprule
\multirow{2}{*}{\textbf{Model}} & \multicolumn{4}{c}{\textbf{Critic}} & \multicolumn{2}{c}{\textbf{Refine}} & \multicolumn{3}{c}{\textbf{Majority Vote}} \\ \cmidrule(lr){2-5} \cmidrule(lr){6-7}  \cmidrule(lr){8-10} 
 & \textbf{P} & \textbf{R} & \textbf{F1} & \textbf{Acc.} & \textbf{Init. Acc} & \textbf{Ref. Acc.} & \textbf{Pass1@N} & \textbf{Maj1@N} & \textbf{+Critic} \\ \midrule
Outcome Label & \textbf{84.4} & 39.0 & 53.3 & 63.0 & 51.8 & 53.6 & 84.0 & 56.2 & 56.2 \\
Process Label & 80.2 & 35.9 & 49.6 & 63.8 & 50.4 & 52.6 & 78.6 & 49.4 & 50.8 \\ \midrule
Only Refine & 62.3 & 60.1 & 61.2 & 66.0 & 55.4 & 49.8 & \textbf{90.4} & 63.0 & 62.8 \\
Only Critic & 67.9 & 75.4 & 71.5 & 71.6 & 52.8 & 55.8 & 89.0 & 60.6 & 60.6 \\ \midrule
Stage 1 & 64.6 & \textbf{93.7} & \textbf{76.5} & 69.0 & 53.2 & 41.2 & \textbf{90.4} & 63.4 & 63.0 \\
Stage 2 & 79.7 & 45.8 & 58.2 & 71.8 & 57.2 & 57.4 & \textbf{90.4} & \textbf{64.6} & 65.0 \\ \midrule
\textbf{Critic-CoT} & 66.1 & 73.7 & 69.7 & \textbf{72.2} & \textbf{57.6} & \textbf{57.8} & 89.2 & \textbf{64.6} & \textbf{66.6} \\ \bottomrule
\end{tabularx}
\caption{MATH500}
\label{table:Ablation Study on MATH500}
\end{subtable}
\caption{Ablation Study on GSM8K and MATH500. We use the metrics from three aspects: critic, including precision, recall, f1-score and accuracy; Iterative Refine, including accuracy before and after the refinement; and Critic As Filter, including Pass1@96, Maj1@96, and Critic+Maj1@96. The ablation study demonstrates the effectiveness of our Critic-CoT design.}
\label{table:Ablation Study Combined}
\end{table*}

\subsection{Critic-CoT Strengthens Out-of-Domain Reasoning}
\label{section:Out-of-Domain Results}

The results of out-domain evaluation are shown in Table~\ref{tab:OOD solution}. In general, our model, especially trained on MATH dataset, achieves positive performance gain with Iteartive Refinement and Critic As Filter, which demonstrates \textbf{the generalized critique and refinement abilities beyond the math training domains}. 
For StrategyQA, our critic models trained on two datasets show a positive performance increase when applying iterative refine and majority vote with the critic filter. On the more challenging dataset AGIEval, 
the Critic-CoT model trained on MATH performs much better than the model trained on grade-school level GSM8K dataset, and shows significant improvements in iterative refinement, rising to 63.7\% compared with 56.6\% prior the Critic-CoT training. On the code generation task HumanEval, the majority vote method is not applicable, but we can still observe the positive improvement with iterative refinement: while the base model's pass rate is 76.2\%, our model trained on GSM8K and MATH achieved 78.1\% and 84.8\% respectively. 
These results highlight the robustness of our Critic-CoT models. 

\section{Ablation Analysis}
\label{section:Ablation Study}
To demonstrate the effectiveness of our Critic-CoT designs, we conduct a series of manual examination and ablation studies, which confirm that our training data is still in high quality (Section \ref{section:The Quality of Constructed Critic-Refine Data}), our proposed stepwise CoT critique is advantageous (Section \ref{section:The Necessity of CoT Critique}), the composition of training data is beneficial (Section \ref{section:The Impact of Training Data}), and the source of improvement shall be attributed to our Critic-CoT framework rather than distillation (Section \ref{section:Comparison with Solution Distilation}).




\subsection{The Quality of Constructed Data}
\label{section:The Quality of Constructed Critic-Refine Data}
To examine the correctness of constructed Critic-CoT training data, we perform the manual evaluation. We sample 300 entries (100 with the critique of correct answer and 100 with the critique and refinemnt of wrong answer) each from the critic-cot data on GSM8K and MATH, and conduct a manual verification to verify the accuracy of the step-wise critiques. For the critique of the correct answer attempt, it is valid if there is indeed no error in all the intermediate steps; for the critique of the wrong attempt, it is valid if the first error step and the reason for the error are both identified. A refinement is correct, if the continuation steps are flawless.

The results of manual verification are demonstrated in Table \ref{table:Human Evaluation on the Critic-CoT Data}, with about 85\% accuracy on wrong-answer critique, and more than 90\% on refinement and correct-answer critique. \textbf{Therefore, the data we automatically constructed maintain a high level of accuracy at the step level, which can well support the critique training process}.

\subsection{The Necessity of CoT Critique}
\label{section:The Necessity of CoT Critique}
To assess the necessity of the stepwise CoT critic, we remove CoT and train two baselines, namely ``Process Label'' and ``Outcome Label''. Specifically, for Process Label, the model is trained to directly predict the correctness of each step, e.g., ``Step 1 is correct. Step 2 is incorrect.''. Further, for Outcome Label, we remove the stepwise labels, which is to predict whether the entire solution is correct without providing step-wise details, e.g., ``Each step from Step 1 to Step 8 is correct'' or ``Some step from Step 1 to Step 8 is incorrect''. These two baselines can be viewed as the natural language versions of the Process Reward Model and Outcome Reward Model.

The results are shown in Table \ref{table:Ablation Study Combined}. We find that removing the Chain-of-Thought intermediate analysis and further stepwise labels, negatively impacts the critic accuracy, which indicates \textbf{the effectiveness of stepwise CoT critique in improving the discrimination performance}. From the original Critic-CoT model to the Process Label Model and the Outcome Label Model, the critique accuracy drops from 92.3\% to 89.5\% and further to 88.0\% on GSM8K, and from 72.2\% to 63.8\% and further to 63.0\% on MATH500. They fail to detect more errors and the recall metric is lower, despite its tendency to more easily pass correct solutions.
Compared with System-1 like reasoning without explicit analysis, System-2 like reasoning with CoT critique can more precisely identify the errors.

\subsection{The Impact of Training Data}
\label{section:The Impact of Training Data}
To evaluate the impact of different data types during training, we perform a vertical ablation by removing either the critic data or the refinement data across both stages. In addition, we perform a horizontal ablation by training only with either stage 1 data or stage 2 data.

As the result shown in Table \ref{table:Ablation Study Combined}, we find that \textbf{training only on the single-type data is less optimal}. 
If only training to refine, the base model's critic ability remains weak, and still \textit{cannot effectively self-correct}, as on GSM8K the accuracy drops from 92.0\% to 88.2\% after iterative refinement, and on MATH500 it drops from 55.4\% to 49.8\%. 
If only training to critic, the model obtains positive improvement from iterative refinement, with increases from 91.2\% to 91.4\% on GSM8K and 52.8\% to 55.8\% on MATH500, but still lags behind to Critic-CoT model that jointly trains the two capabilities. 
And finally, in terms of critic and reasoning performance, combining the data from two stages is better than training the data only from one stage.

\begin{table}[h]
\centering
\small
\begin{tabularx}{0.48\textwidth}{@{}lll@{}}
\toprule
\textbf{Model} & \textbf{GSM8K} & \textbf{MATH500} \\ \midrule
Llama-3-70B-Instruct & 89.6 & 50.4 \\
GPT4-Turbo Solution & 90.7 & 48.0 \\
Critic-CoT & \textbf{93.3} & \textbf{57.8} \\ \bottomrule
\end{tabularx}
\caption{Comparision between solution distillation and our Critic-CoT model with Iterative Refine, with the metric of Top-1 Accuracy. Directly training on the trajectories of advanced LLMs did not necessarily improve the performance.}

\label{table:distilation}

\end{table}


\subsection{Comparison with Solution Distilation} 
\label{section:Comparison with Solution Distilation}
Since our data construction process leverages GPT-4 Turbo, one potential concern is whether the performance increase comes from distilling GPT-4, or our proposed Critic-CoT framework.
To address this concern and better understand the source of improvement, we also train the traditional rejection sampling fine-tuning (RFT) baseline, using GPT4-Turbo to generate a correct-answer solution for each problem on the dataset of GSM8K and MATH respectively. 

As the result shown in Table \ref{table:distilation}, unlike Critic-CoT, \textbf{directly distilling from frontier LLMs may not significantly improve the reasoning performance}, with only 90.7\% on GSM8K and 48.0\% on MAT500, which is less superior. On the one hand, the Llama-3-70B model has already experienced heavy post-training for downstream reasoning tasks \citep{llama3.1modelcard}. On the other hand, as shown in Section \ref{section:Main Results}, the improvements can be attributed to two key factors. First, strengthening the ability to critique and refinement, which is under-trained, directly improves the Top-1 accuracy. Moreover, at the inference phase, we can actively leverage the model’s ability to reflect on its reasoning and correct mistakes, thus obtaining additional improvements that are not applicable without critic-refine training.
\textbf{Therefore, our proposed Critic-CoT framework plays a crucial role in driving the observed performance improvements.}

\section{Conclusion}

In this paper, we introduced the Critic-CoT paradigm to enhance the reasoning abilities of Large Language Models, through a more System-2-like, step-by-step Chain-of-Thought critique. Our approach leverages weak supervision to construct training data for critiques and refinements, thereby reducing the reliance on extensive human annotation. We demonstrated the effectiveness of our method through substantial improvements across the dataset of GSM8K and MATH. Additionally, our results present that training on the capabilities of critique and refinement alone improves task-solving performance, which indicates a mutual-reinforce mechanism within the LLMs. We hope our work may inspire further investigations into the advancement of the self-critic framework and the transition toward System-2 reasoning.

\section*{Limitations}
In this paper, we propose Critic-CoT, a framework to automatically construct critic-refine training data in the reasoning domain, where the correctness of a solution is verifiable by checking the final answer. For more general tasks, how to filter valid critiques and train robust critic models is worth investigating in future works.

\section*{Ethics Statement}
All the data and models are acquired from public datasets and pre-trained models, and no human annotators are involved during the data construction procedure.

\section*{Acknowledgment}
We sincerely thank the reviewers for their insightful comments and valuable suggestions. This work was supported by Beijing Natural Science Foundation (L243006), Beijing Municipal Science and Technology Project (Nos. Z231100010323002), and the Natural Science Foundation of China (No. 62476265, 62306303).

\bibliography{acl_latex}

\begin{thebibliography}{68}
\providecommand{\natexlab}[1]{#1}

\bibitem[{AI@Meta(2024)}]{llama3.1modelcard}
AI@Meta. 2024.
\newblock \href {https://github.com/meta-llama/llama-models/blob/main/models/llama3_1/MODEL_CARD.md} {Llama 3.1 model card}.

\bibitem[{Asai et~al.(2024)Asai, Wu, Wang, Sil, and Hajishirzi}]{asai2024selfrag}
Akari Asai, Zeqiu Wu, Yizhong Wang, Avirup Sil, and Hannaneh Hajishirzi. 2024.
\newblock \href {https://openreview.net/forum?id=hSyW5go0v8} {Self-{RAG}: Learning to retrieve, generate, and critique through self-reflection}.
\newblock In \emph{The Twelfth International Conference on Learning Representations}.

\bibitem[{Bubeck et~al.(2023)Bubeck, Chandrasekaran, Eldan, Gehrke, Horvitz, Kamar, Lee, Lee, Li, Lundberg, Nori, Palangi, Ribeiro, and Zhang}]{bubeck2023sparksartificialgeneralintelligence}
Sébastien Bubeck, Varun Chandrasekaran, Ronen Eldan, Johannes Gehrke, Eric Horvitz, Ece Kamar, Peter Lee, Yin~Tat Lee, Yuanzhi Li, Scott Lundberg, Harsha Nori, Hamid Palangi, Marco~Tulio Ribeiro, and Yi~Zhang. 2023.
\newblock \href {https://arxiv.org/abs/2303.12712} {Sparks of artificial general intelligence: Early experiments with gpt-4}.
\newblock \emph{Preprint}, arXiv:2303.12712.

\bibitem[{Burns et~al.(2023)Burns, Izmailov, Kirchner, Baker, Gao, Aschenbrenner, Chen, Ecoffet, Joglekar, Leike, Sutskever, and Wu}]{burns2023weaktostronggeneralizationelicitingstrong}
Collin Burns, Pavel Izmailov, Jan~Hendrik Kirchner, Bowen Baker, Leo Gao, Leopold Aschenbrenner, Yining Chen, Adrien Ecoffet, Manas Joglekar, Jan Leike, Ilya Sutskever, and Jeff Wu. 2023.
\newblock \href {https://arxiv.org/abs/2312.09390} {Weak-to-strong generalization: Eliciting strong capabilities with weak supervision}.
\newblock \emph{Preprint}, arXiv:2312.09390.

\bibitem[{Cao et~al.(2024)Cao, Lu, Lu, Chen, Ren, Xiang, Liu, Lu, He, Han, Sun, Lin, and Yu}]{cao2024scalableautomatedalignmentllms}
Boxi Cao, Keming Lu, Xinyu Lu, Jiawei Chen, Mengjie Ren, Hao Xiang, Peilin Liu, Yaojie Lu, Ben He, Xianpei Han, Le~Sun, Hongyu Lin, and Bowen Yu. 2024.
\newblock \href {https://arxiv.org/abs/2406.01252} {Towards scalable automated alignment of llms: A survey}.
\newblock \emph{Preprint}, arXiv:2406.01252.

\bibitem[{Chan et~al.(2024)Chan, Chen, Su, Yu, Xue, Zhang, Fu, and Liu}]{chan2024chateval}
Chi-Min Chan, Weize Chen, Yusheng Su, Jianxuan Yu, Wei Xue, Shanghang Zhang, Jie Fu, and Zhiyuan Liu. 2024.
\newblock \href {https://openreview.net/forum?id=FQepisCUWu} {Chateval: Towards better {LLM}-based evaluators through multi-agent debate}.
\newblock In \emph{The Twelfth International Conference on Learning Representations}.

\bibitem[{Chen et~al.(2024{\natexlab{a}})Chen, Liao, Li, and Fan}]{DBLP:journals/corr/abs-2405-03553}
Guoxin Chen, Minpeng Liao, Chengxi Li, and Kai Fan. 2024{\natexlab{a}}.
\newblock \href {https://doi.org/10.48550/ARXIV.2405.03553} {Alphamath almost zero: process supervision without process}.
\newblock \emph{CoRR}, abs/2405.03553.

\bibitem[{Chen et~al.(2021)Chen, Tworek, Jun, Yuan, de~Oliveira~Pinto, Kaplan, Edwards, Burda, Joseph, Brockman, Ray, Puri, Krueger, Petrov, Khlaaf, Sastry, Mishkin, Chan, Gray, Ryder, Pavlov, Power, Kaiser, Bavarian, Winter, Tillet, Such, Cummings, Plappert, Chantzis, Barnes, Herbert-Voss, Guss, Nichol, Paino, Tezak, Tang, Babuschkin, Balaji, Jain, Saunders, Hesse, Carr, Leike, Achiam, Misra, Morikawa, Radford, Knight, Brundage, Murati, Mayer, Welinder, McGrew, Amodei, McCandlish, Sutskever, and Zaremba}]{chen2021evaluatinglargelanguagemodels}
Mark Chen, Jerry Tworek, Heewoo Jun, Qiming Yuan, Henrique~Ponde de~Oliveira~Pinto, Jared Kaplan, Harri Edwards, Yuri Burda, Nicholas Joseph, Greg Brockman, Alex Ray, Raul Puri, Gretchen Krueger, Michael Petrov, Heidy Khlaaf, Girish Sastry, Pamela Mishkin, Brooke Chan, Scott Gray, Nick Ryder, Mikhail Pavlov, Alethea Power, Lukasz Kaiser, Mohammad Bavarian, Clemens Winter, Philippe Tillet, Felipe~Petroski Such, Dave Cummings, Matthias Plappert, Fotios Chantzis, Elizabeth Barnes, Ariel Herbert-Voss, William~Hebgen Guss, Alex Nichol, Alex Paino, Nikolas Tezak, Jie Tang, Igor Babuschkin, Suchir Balaji, Shantanu Jain, William Saunders, Christopher Hesse, Andrew~N. Carr, Jan Leike, Josh Achiam, Vedant Misra, Evan Morikawa, Alec Radford, Matthew Knight, Miles Brundage, Mira Murati, Katie Mayer, Peter Welinder, Bob McGrew, Dario Amodei, Sam McCandlish, Ilya Sutskever, and Wojciech Zaremba. 2021.
\newblock \href {https://arxiv.org/abs/2107.03374} {Evaluating large language models trained on code}.
\newblock \emph{Preprint}, arXiv:2107.03374.

\bibitem[{Chen et~al.(2024{\natexlab{b}})Chen, Xu, Liang, He, Pang, Yu, Song, Liu, Zhou, Zhang et~al.}]{chen2024not}
Xingyu Chen, Jiahao Xu, Tian Liang, Zhiwei He, Jianhui Pang, Dian Yu, Linfeng Song, Qiuzhi Liu, Mengfei Zhou, Zhuosheng Zhang, et~al. 2024{\natexlab{b}}.
\newblock Do not think that much for 2+ 3=? on the overthinking of o1-like llms.
\newblock \emph{arXiv preprint arXiv:2412.21187}.

\bibitem[{Chollet(2019)}]{chollet2019measureintelligence}
François Chollet. 2019.
\newblock \href {https://arxiv.org/abs/1911.01547} {On the measure of intelligence}.
\newblock \emph{Preprint}, arXiv:1911.01547.

\bibitem[{Cobbe et~al.(2021{\natexlab{a}})Cobbe, Kosaraju, Bavarian, Chen, Jun, Kaiser, Plappert, Tworek, Hilton, Nakano, Hesse, and Schulman}]{DBLP:journals/corr/abs-2110-14168}
Karl Cobbe, Vineet Kosaraju, Mohammad Bavarian, Mark Chen, Heewoo Jun, Lukasz Kaiser, Matthias Plappert, Jerry Tworek, Jacob Hilton, Reiichiro Nakano, Christopher Hesse, and John Schulman. 2021{\natexlab{a}}.
\newblock \href {https://arxiv.org/abs/2110.14168} {Training verifiers to solve math word problems}.
\newblock \emph{CoRR}, abs/2110.14168.

\bibitem[{Cobbe et~al.(2021{\natexlab{b}})Cobbe, Kosaraju, Bavarian, Chen, Jun, Kaiser, Plappert, Tworek, Hilton, Nakano, Hesse, and Schulman}]{cobbe2021trainingverifierssolvemath}
Karl Cobbe, Vineet Kosaraju, Mohammad Bavarian, Mark Chen, Heewoo Jun, Lukasz Kaiser, Matthias Plappert, Jerry Tworek, Jacob Hilton, Reiichiro Nakano, Christopher Hesse, and John Schulman. 2021{\natexlab{b}}.
\newblock \href {https://arxiv.org/abs/2110.14168} {Training verifiers to solve math word problems}.
\newblock \emph{Preprint}, arXiv:2110.14168.

\bibitem[{DeepSeek{-}AI et~al.(2024)DeepSeek{-}AI, Liu, Feng, Wang, Wang, Liu, and et~al.}]{DBLP:journals/corr/abs-2405-04434}
DeepSeek{-}AI, Aixin Liu, Bei Feng, Bin Wang, Bingxuan Wang, Bo~Liu, and et~al. 2024.
\newblock \href {https://doi.org/10.48550/ARXIV.2405.04434} {Deepseek-v2: {A} strong, economical, and efficient mixture-of-experts language model}.
\newblock \emph{CoRR}, abs/2405.04434.

\bibitem[{Dubey et~al.(2024)Dubey, Jauhri, Pandey, Kadian, Al-Dahle, Letman, Mathur, Schelten, Yang, Fan, Goyal, Hartshorn, Yang, Mitra, and et~al.}]{llama3modelcard}
Abhimanyu Dubey, Abhinav Jauhri, Abhinav Pandey, Abhishek Kadian, Ahmad Al-Dahle, Aiesha Letman, Akhil Mathur, Alan Schelten, Amy Yang, Angela Fan, Anirudh Goyal, Anthony Hartshorn, Aobo Yang, Archi Mitra, and et~al. 2024.
\newblock \href {https://arxiv.org/abs/2407.21783} {The llama 3 herd of models}.
\newblock \emph{Preprint}, arXiv:2407.21783.

\bibitem[{Feng et~al.(2023)Feng, Zhang, Gu, Ye, He, and Wang}]{feng2023towards}
Guhao Feng, Bohang Zhang, Yuntian Gu, Haotian Ye, Di~He, and Liwei Wang. 2023.
\newblock \href {https://openreview.net/forum?id=qHrADgAdYu} {Towards revealing the mystery behind chain of thought: A theoretical perspective}.
\newblock In \emph{Thirty-seventh Conference on Neural Information Processing Systems}.

\bibitem[{Gandhi et~al.(2025)Gandhi, Chakravarthy, Singh, Lile, and Goodman}]{gandhi2025cognitive}
Kanishk Gandhi, Ayush Chakravarthy, Anikait Singh, Nathan Lile, and Noah~D Goodman. 2025.
\newblock Cognitive behaviors that enable self-improving reasoners, or, four habits of highly effective stars.
\newblock \emph{arXiv preprint arXiv:2503.01307}.

\bibitem[{Gao et~al.(2024)Gao, Cai, Xu, Wang, Zheng, Lin, Lu, Lin, Zhou, Xiao, Hu, Liu, and Chang}]{DBLP:journals/corr/abs-2406-14024}
Bofei Gao, Zefan Cai, Runxin Xu, Peiyi Wang, Ce~Zheng, Runji Lin, Keming Lu, Junyang Lin, Chang Zhou, Wen Xiao, Junjie Hu, Tianyu Liu, and Baobao Chang. 2024.
\newblock \href {https://doi.org/10.48550/ARXIV.2406.14024} {{LLM} critics help catch bugs in mathematics: Towards a better mathematical verifier with natural language feedback}.
\newblock \emph{CoRR}, abs/2406.14024.

\bibitem[{Geva et~al.(2021)Geva, Khashabi, Segal, Khot, Roth, and Berant}]{geva-etal-2021-aristotle}
Mor Geva, Daniel Khashabi, Elad Segal, Tushar Khot, Dan Roth, and Jonathan Berant. 2021.
\newblock \href {https://doi.org/10.1162/tacl_a_00370} {Did aristotle use a laptop? a question answering benchmark with implicit reasoning strategies}.
\newblock \emph{Transactions of the Association for Computational Linguistics}, 9:346--361.

\bibitem[{Guo et~al.(2025)Guo, Yang, Zhang, Song, Zhang, Xu, Zhu, Ma, Wang, Bi et~al.}]{guo2025deepseek}
Daya Guo, Dejian Yang, Haowei Zhang, Junxiao Song, Ruoyu Zhang, Runxin Xu, Qihao Zhu, Shirong Ma, Peiyi Wang, Xiao Bi, et~al. 2025.
\newblock Deepseek-r1: Incentivizing reasoning capability in llms via reinforcement learning.
\newblock \emph{arXiv preprint arXiv:2501.12948}.

\bibitem[{Havrilla et~al.(2024)Havrilla, Du, Raparthy, Nalmpantis, Dwivedi-Yu, Hambro, Sukhbaatar, and Raileanu}]{havrilla2024teaching}
Alexander Havrilla, Yuqing Du, Sharath~Chandra Raparthy, Christoforos Nalmpantis, Jane Dwivedi-Yu, Eric Hambro, Sainbayar Sukhbaatar, and Roberta Raileanu. 2024.
\newblock \href {https://openreview.net/forum?id=mjqoceuMnI} {Teaching large language models to reason with reinforcement learning}.
\newblock In \emph{AI for Math Workshop @ ICML 2024}.

\bibitem[{He et~al.(2025)He, Li, Liu, Wang, Bu, Zhang, Peng, Zhang, Su, and Zheng}]{he2025can}
Yancheng He, Shilong Li, Jiaheng Liu, Weixun Wang, Xingyuan Bu, Ge~Zhang, Zhongyuan Peng, Zhaoxiang Zhang, Wenbo Su, and Bo~Zheng. 2025.
\newblock Can large language models detect errors in long chain-of-thought reasoning?
\newblock \emph{arXiv preprint arXiv:2502.19361}.

\bibitem[{Hegel et~al.(1991)Hegel, Geraets, Suchting, and Harris}]{hegel1991encyclopaedia}
G.W.F. Hegel, T.F. Geraets, W.A. Suchting, and H.S. Harris. 1991.
\newblock \href {https://books.google.ca/books?id=4BNUFZ_hQ1wC} {\emph{The Encyclopaedia Logic, with the Zustze: Part I of the Encyclopaedia of Philosophical Sciences with the Zusatze}}.
\newblock Hackett Classics Series. Hackett.

\bibitem[{Hendrycks et~al.(2021)Hendrycks, Burns, Kadavath, Arora, Basart, Tang, Song, and Steinhardt}]{hendrycks2021measuring}
Dan Hendrycks, Collin Burns, Saurav Kadavath, Akul Arora, Steven Basart, Eric Tang, Dawn Song, and Jacob Steinhardt. 2021.
\newblock \href {https://openreview.net/forum?id=7Bywt2mQsCe} {Measuring mathematical problem solving with the {MATH} dataset}.
\newblock In \emph{Thirty-fifth Conference on Neural Information Processing Systems Datasets and Benchmarks Track (Round 2)}.

\bibitem[{Huang et~al.(2024)Huang, Chen, Mishra, Zheng, Yu, Song, and Zhou}]{huang2024large}
Jie Huang, Xinyun Chen, Swaroop Mishra, Huaixiu~Steven Zheng, Adams~Wei Yu, Xinying Song, and Denny Zhou. 2024.
\newblock \href {https://openreview.net/forum?id=IkmD3fKBPQ} {Large language models cannot self-correct reasoning yet}.
\newblock In \emph{The Twelfth International Conference on Learning Representations}.

\bibitem[{Ji et~al.(2023)Ji, Yu, Xu, Lee, Ishii, and Fung}]{ji-etal-2023-towards}
Ziwei Ji, Tiezheng Yu, Yan Xu, Nayeon Lee, Etsuko Ishii, and Pascale Fung. 2023.
\newblock \href {https://doi.org/10.18653/v1/2023.findings-emnlp.123} {Towards mitigating {LLM} hallucination via self reflection}.
\newblock In \emph{Findings of the Association for Computational Linguistics: EMNLP 2023}, pages 1827--1843, Singapore. Association for Computational Linguistics.

\bibitem[{Kahneman(2011)}]{kahneman2011thinking}
Daniel Kahneman. 2011.
\newblock \emph{Thinking Fast and Slow}.
\newblock Farrar, Straus and Giroux.

\bibitem[{Kierkegaard(1989)}]{d0f1f802-4f88-3f86-ab42-b3525806553e}
Søren Kierkegaard. 1989.
\newblock \href {http://www.jstor.org/stable/j.ctt24hr3n} {\emph{Kierkegaard's Writings, II, Volume 2: The Concept of Irony, with Continual Reference to Socrates/Notes of Schelling's Berlin Lectures}}.
\newblock Princeton University Press.

\bibitem[{Kim et~al.(2023)Kim, Baldi, and McAleer}]{kim2023language}
Geunwoo Kim, Pierre Baldi, and Stephen~Marcus McAleer. 2023.
\newblock \href {https://openreview.net/forum?id=M6OmjAZ4CX} {Language models can solve computer tasks}.
\newblock In \emph{Thirty-seventh Conference on Neural Information Processing Systems}.

\bibitem[{Kwon et~al.(2023)Kwon, Li, Zhuang, Sheng, Zheng, Yu, Gonzalez, Zhang, and Stoica}]{10.1145/3600006.3613165}
Woosuk Kwon, Zhuohan Li, Siyuan Zhuang, Ying Sheng, Lianmin Zheng, Cody~Hao Yu, Joseph Gonzalez, Hao Zhang, and Ion Stoica. 2023.
\newblock \href {https://doi.org/10.1145/3600006.3613165} {Efficient memory management for large language model serving with pagedattention}.
\newblock In \emph{Proceedings of the 29th Symposium on Operating Systems Principles}, SOSP '23, page 611–626, New York, NY, USA. Association for Computing Machinery.

\bibitem[{Leike et~al.(2018)Leike, Krueger, Everitt, Martic, Maini, and Legg}]{leike2018scalableagentalignmentreward}
Jan Leike, David Krueger, Tom Everitt, Miljan Martic, Vishal Maini, and Shane Legg. 2018.
\newblock \href {https://arxiv.org/abs/1811.07871} {Scalable agent alignment via reward modeling: a research direction}.
\newblock \emph{Preprint}, arXiv:1811.07871.

\bibitem[{Li et~al.(2024)Li, Liu, Zhou, and Ma}]{li2024chain}
Zhiyuan Li, Hong Liu, Denny Zhou, and Tengyu Ma. 2024.
\newblock \href {https://openreview.net/forum?id=3EWTEy9MTM} {Chain of thought empowers transformers to solve inherently serial problems}.
\newblock In \emph{The Twelfth International Conference on Learning Representations}.

\bibitem[{Liang et~al.(2024)Liang, He, Jiao, Wang, Wang, Wang, Yang, Shi, and Tu}]{liang-etal-2024-encouraging}
Tian Liang, Zhiwei He, Wenxiang Jiao, Xing Wang, Yan Wang, Rui Wang, Yujiu Yang, Shuming Shi, and Zhaopeng Tu. 2024.
\newblock \href {https://doi.org/10.18653/v1/2024.emnlp-main.992} {Encouraging divergent thinking in large language models through multi-agent debate}.
\newblock In \emph{Proceedings of the 2024 Conference on Empirical Methods in Natural Language Processing}, pages 17889--17904, Miami, Florida, USA. Association for Computational Linguistics.

\bibitem[{Lightman et~al.(2024)Lightman, Kosaraju, Burda, Edwards, Baker, Lee, Leike, Schulman, Sutskever, and Cobbe}]{lightman2024lets}
Hunter Lightman, Vineet Kosaraju, Yuri Burda, Harrison Edwards, Bowen Baker, Teddy Lee, Jan Leike, John Schulman, Ilya Sutskever, and Karl Cobbe. 2024.
\newblock \href {https://openreview.net/forum?id=v8L0pN6EOi} {Let's verify step by step}.
\newblock In \emph{The Twelfth International Conference on Learning Representations}.

\bibitem[{Lin et~al.(2024)Lin, Gou, Liang, Luo, Liu, and Yang}]{lin-etal-2024-criticbench}
Zicheng Lin, Zhibin Gou, Tian Liang, Ruilin Luo, Haowei Liu, and Yujiu Yang. 2024.
\newblock \href {https://doi.org/10.18653/v1/2024.findings-acl.91} {{C}ritic{B}ench: Benchmarking {LLM}s for critique-correct reasoning}.
\newblock In \emph{Findings of the Association for Computational Linguistics: ACL 2024}, pages 1552--1587, Bangkok, Thailand. Association for Computational Linguistics.

\bibitem[{Liu et~al.(2023)Liu, Singh, Freeman, Co{-}Reyes, and Liu}]{DBLP:journals/corr/abs-2310-10047}
Yixin Liu, Avi Singh, C.~Daniel Freeman, John~D. Co{-}Reyes, and Peter~J. Liu. 2023.
\newblock \href {https://doi.org/10.48550/ARXIV.2310.10047} {Improving large language model fine-tuning for solving math problems}.
\newblock \emph{CoRR}, abs/2310.10047.

\bibitem[{Luo et~al.(2023)Luo, Lin, Liu, Shu, Zhu, Shang, and Meng}]{luo2023critiqueabilitylargelanguage}
Liangchen Luo, Zi~Lin, Yinxiao Liu, Lei Shu, Yun Zhu, Jingbo Shang, and Lei Meng. 2023.
\newblock \href {https://arxiv.org/abs/2310.04815} {Critique ability of large language models}.
\newblock \emph{Preprint}, arXiv:2310.04815.

\bibitem[{Luo et~al.(2024)Luo, Liu, Liu, Phatale, Lara, Li, Shu, Zhu, Meng, Sun, and Rastogi}]{luo2024improvemathematicalreasoninglanguage}
Liangchen Luo, Yinxiao Liu, Rosanne Liu, Samrat Phatale, Harsh Lara, Yunxuan Li, Lei Shu, Yun Zhu, Lei Meng, Jiao Sun, and Abhinav Rastogi. 2024.
\newblock \href {https://arxiv.org/abs/2406.06592} {Improve mathematical reasoning in language models by automated process supervision}.
\newblock \emph{Preprint}, arXiv:2406.06592.

\bibitem[{Madaan et~al.(2023)Madaan, Tandon, Gupta, Hallinan, Gao, Wiegreffe, Alon, Dziri, Prabhumoye, Yang, Gupta, Majumder, Hermann, Welleck, Yazdanbakhsh, and Clark}]{madaan2023selfrefine}
Aman Madaan, Niket Tandon, Prakhar Gupta, Skyler Hallinan, Luyu Gao, Sarah Wiegreffe, Uri Alon, Nouha Dziri, Shrimai Prabhumoye, Yiming Yang, Shashank Gupta, Bodhisattwa~Prasad Majumder, Katherine Hermann, Sean Welleck, Amir Yazdanbakhsh, and Peter Clark. 2023.
\newblock \href {https://openreview.net/forum?id=S37hOerQLB} {Self-refine: Iterative refinement with self-feedback}.
\newblock In \emph{Thirty-seventh Conference on Neural Information Processing Systems}.

\bibitem[{McAleese et~al.(2024)McAleese, Pokorny, Uribe, Nitishinskaya, Trebacz, and Leike}]{mcaleese2024llmcriticshelpcatch}
Nat McAleese, Rai~Michael Pokorny, Juan Felipe~Ceron Uribe, Evgenia Nitishinskaya, Maja Trebacz, and Jan Leike. 2024.
\newblock \href {https://arxiv.org/abs/2407.00215} {Llm critics help catch llm bugs}.
\newblock \emph{Preprint}, arXiv:2407.00215.

\bibitem[{Morris et~al.(2024)Morris, Sohl-Dickstein, Fiedel, Warkentin, Dafoe, Faust, Farabet, and Legg}]{morris2024position}
Meredith~Ringel Morris, Jascha Sohl-Dickstein, Noah Fiedel, Tris Warkentin, Allan Dafoe, Aleksandra Faust, Clement Farabet, and Shane Legg. 2024.
\newblock \href {https://openreview.net/forum?id=0ofzEysK2D} {Position: Levels of {AGI} for operationalizing progress on the path to {AGI}}.
\newblock In \emph{Forty-first International Conference on Machine Learning}.

\bibitem[{OpenAI(2023)}]{DBLP:journals/corr/abs-2303-08774}
OpenAI. 2023.
\newblock \href {https://doi.org/10.48550/ARXIV.2303.08774} {{GPT-4} technical report}.
\newblock \emph{CoRR}, abs/2303.08774.

\bibitem[{OpenAI(2024)}]{openaio1}
OpenAI. 2024.
\newblock \href {https://assets.ctfassets.net/kftzwdyauwt9/67qJD51Aur3eIc96iOfeOP/71551c3d223cd97e591aa89567306912/o1_system_card.pdf} {Openai o1 system card}.

\bibitem[{Ouyang et~al.(2022)Ouyang, Wu, Jiang, Almeida, Wainwright, Mishkin, Zhang, Agarwal, Slama, Ray, Schulman, Hilton, Kelton, Miller, Simens, Askell, Welinder, Christiano, Leike, and Lowe}]{NEURIPS2022_b1efde53}
Long Ouyang, Jeffrey Wu, Xu~Jiang, Diogo Almeida, Carroll Wainwright, Pamela Mishkin, Chong Zhang, Sandhini Agarwal, Katarina Slama, Alex Ray, John Schulman, Jacob Hilton, Fraser Kelton, Luke Miller, Maddie Simens, Amanda Askell, Peter Welinder, Paul~F Christiano, Jan Leike, and Ryan Lowe. 2022.
\newblock \href {https://proceedings.neurips.cc/paper_files/paper/2022/file/b1efde53be364a73914f58805a001731-Paper-Conference.pdf} {Training language models to follow instructions with human feedback}.
\newblock In \emph{Advances in Neural Information Processing Systems}, volume~35, pages 27730--27744. Curran Associates, Inc.

\bibitem[{Phan et~al.(2025)Phan, Gatti, Han, Li, Hu, Zhang, Zhang, Shaaban, Ling, Shi et~al.}]{phan2025humanity}
Long Phan, Alice Gatti, Ziwen Han, Nathaniel Li, Josephina Hu, Hugh Zhang, Chen Bo~Calvin Zhang, Mohamed Shaaban, John Ling, Sean Shi, et~al. 2025.
\newblock Humanity's last exam.
\newblock \emph{arXiv preprint arXiv:2501.14249}.

\bibitem[{Popper(1934)}]{Popper1934-POPTLO-2}
Karl~Raimund Popper. 1934.
\newblock \emph{The Logic of Scientific Discovery}.
\newblock Routledge, New York.

\bibitem[{Rajbhandari et~al.(2021)Rajbhandari, Ruwase, Rasley, Smith, and He}]{10.1145/3458817.3476205}
Samyam Rajbhandari, Olatunji Ruwase, Jeff Rasley, Shaden Smith, and Yuxiong He. 2021.
\newblock \href {https://doi.org/10.1145/3458817.3476205} {Zero-infinity: breaking the gpu memory wall for extreme scale deep learning}.
\newblock In \emph{Proceedings of the International Conference for High Performance Computing, Networking, Storage and Analysis}, SC '21, New York, NY, USA. Association for Computing Machinery.

\bibitem[{Saunders et~al.(2022)Saunders, Yeh, Wu, Bills, Ouyang, Ward, and Leike}]{saunders2022selfcritiquingmodelsassistinghuman}
William Saunders, Catherine Yeh, Jeff Wu, Steven Bills, Long Ouyang, Jonathan Ward, and Jan Leike. 2022.
\newblock \href {https://arxiv.org/abs/2206.05802} {Self-critiquing models for assisting human evaluators}.
\newblock \emph{Preprint}, arXiv:2206.05802.

\bibitem[{Schaul(2024)}]{schaul2024boundless}
Tom Schaul. 2024.
\newblock Boundless socratic learning with language games.
\newblock \emph{arXiv preprint arXiv:2411.16905}.

\bibitem[{Shinn et~al.(2023)Shinn, Cassano, Gopinath, Narasimhan, and Yao}]{shinn2023reflexion}
Noah Shinn, Federico Cassano, Ashwin Gopinath, Karthik~R Narasimhan, and Shunyu Yao. 2023.
\newblock \href {https://openreview.net/forum?id=vAElhFcKW6} {Reflexion: language agents with verbal reinforcement learning}.
\newblock In \emph{Thirty-seventh Conference on Neural Information Processing Systems}.

\bibitem[{Snell et~al.(2024)Snell, Lee, Xu, and Kumar}]{snell2024scalingllmtesttimecompute}
Charlie Snell, Jaehoon Lee, Kelvin Xu, and Aviral Kumar. 2024.
\newblock \href {https://arxiv.org/abs/2408.03314} {Scaling llm test-time compute optimally can be more effective than scaling model parameters}.
\newblock \emph{Preprint}, arXiv:2408.03314.

\bibitem[{Tong et~al.(2024)Tong, Zhang, Wang, Wu, and He}]{tong2024dartmathdifficultyawarerejectiontuning}
Yuxuan Tong, Xiwen Zhang, Rui Wang, Ruidong Wu, and Junxian He. 2024.
\newblock \href {https://arxiv.org/abs/2407.13690} {Dart-math: Difficulty-aware rejection tuning for mathematical problem-solving}.
\newblock \emph{Preprint}, arXiv:2407.13690.

\bibitem[{Wang et~al.(2023{\natexlab{a}})Wang, Li, Shao, Xu, Dai, Li, Chen, Wu, and Sui}]{DBLP:journals/corr/abs-2312-08935}
Peiyi Wang, Lei Li, Zhihong Shao, R.~X. Xu, Damai Dai, Yifei Li, Deli Chen, Y.~Wu, and Zhifang Sui. 2023{\natexlab{a}}.
\newblock \href {https://doi.org/10.48550/ARXIV.2312.08935} {Math-shepherd: Verify and reinforce llms step-by-step without human annotations}.
\newblock \emph{CoRR}, abs/2312.08935.

\bibitem[{Wang et~al.(2024)Wang, Kulikov, Golovneva, Yu, Yuan, Dwivedi-Yu, Pang, Fazel-Zarandi, Weston, and Li}]{wang2024selftaughtevaluators}
Tianlu Wang, Ilia Kulikov, Olga Golovneva, Ping Yu, Weizhe Yuan, Jane Dwivedi-Yu, Richard~Yuanzhe Pang, Maryam Fazel-Zarandi, Jason Weston, and Xian Li. 2024.
\newblock \href {https://arxiv.org/abs/2408.02666} {Self-taught evaluators}.
\newblock \emph{Preprint}, arXiv:2408.02666.

\bibitem[{Wang et~al.(2023{\natexlab{b}})Wang, Wei, Schuurmans, Le, Chi, Narang, Chowdhery, and Zhou}]{wang2023selfconsistency}
Xuezhi Wang, Jason Wei, Dale Schuurmans, Quoc~V Le, Ed~H. Chi, Sharan Narang, Aakanksha Chowdhery, and Denny Zhou. 2023{\natexlab{b}}.
\newblock \href {https://openreview.net/forum?id=1PL1NIMMrw} {Self-consistency improves chain of thought reasoning in language models}.
\newblock In \emph{The Eleventh International Conference on Learning Representations}.

\bibitem[{West et~al.(2024)West, Lu, Dziri, Brahman, Li, Hwang, Jiang, Fisher, Ravichander, Chandu, Newman, Koh, Ettinger, and Choi}]{west2024the}
Peter West, Ximing Lu, Nouha Dziri, Faeze Brahman, Linjie Li, Jena~D. Hwang, Liwei Jiang, Jillian Fisher, Abhilasha Ravichander, Khyathi Chandu, Benjamin Newman, Pang~Wei Koh, Allyson Ettinger, and Yejin Choi. 2024.
\newblock \href {https://openreview.net/forum?id=CF8H8MS5P8} {The generative {AI} paradox: {\textquotedblleft}what it can create, it may not understand{\textquotedblright}}.
\newblock In \emph{The Twelfth International Conference on Learning Representations}.

\bibitem[{Wolf et~al.(2020)Wolf, Debut, Sanh, Chaumond, Delangue, Moi, Cistac, Rault, Louf, Funtowicz, Davison, Shleifer, von Platen, Ma, Jernite, Plu, Xu, Scao, Gugger, Drame, Lhoest, and Rush}]{wolf-etal-2020-transformers}
Thomas Wolf, Lysandre Debut, Victor Sanh, Julien Chaumond, Clement Delangue, Anthony Moi, Pierric Cistac, Tim Rault, Rémi Louf, Morgan Funtowicz, Joe Davison, Sam Shleifer, Patrick von Platen, Clara Ma, Yacine Jernite, Julien Plu, Canwen Xu, Teven~Le Scao, Sylvain Gugger, Mariama Drame, Quentin Lhoest, and Alexander~M. Rush. 2020.
\newblock \href {https://www.aclweb.org/anthology/2020.emnlp-demos.6} {Transformers: State-of-the-art natural language processing}.
\newblock In \emph{Proceedings of the 2020 Conference on Empirical Methods in Natural Language Processing: System Demonstrations}, pages 38--45, Online. Association for Computational Linguistics.

\bibitem[{Wu et~al.(2024)Wu, Zeng, Zhang, Tan, Shen, and Jiang}]{wu-etal-2024-large}
Zhenyu Wu, Qingkai Zeng, Zhihan Zhang, Zhaoxuan Tan, Chao Shen, and Meng Jiang. 2024.
\newblock \href {https://aclanthology.org/2024.emnlp-main.714} {Large language models can self-correct with key condition verification}.
\newblock In \emph{Proceedings of the 2024 Conference on Empirical Methods in Natural Language Processing}, pages 12846--12867, Miami, Florida, USA. Association for Computational Linguistics.

\bibitem[{Yan et~al.(2024)Yan, Zhu, Wang, Gui, and He}]{yan-etal-2024-mirror}
Hanqi Yan, Qinglin Zhu, Xinyu Wang, Lin Gui, and Yulan He. 2024.
\newblock \href {https://doi.org/10.18653/v1/2024.acl-long.382} {Mirror: Multiple-perspective self-reflection method for knowledge-rich reasoning}.
\newblock In \emph{Proceedings of the 62nd Annual Meeting of the Association for Computational Linguistics (Volume 1: Long Papers)}, pages 7086--7103, Bangkok, Thailand. Association for Computational Linguistics.

\bibitem[{Yang et~al.(2024)Yang, Yang, Hui, Zheng, Yu, Zhou, and et~al.}]{yang2024qwen2technicalreport}
An~Yang, Baosong Yang, Binyuan Hui, Bo~Zheng, Bowen Yu, Chang Zhou, and et~al. 2024.
\newblock \href {https://arxiv.org/abs/2407.10671} {Qwen2 technical report}.
\newblock \emph{Preprint}, arXiv:2407.10671.

\bibitem[{Ying et~al.(2024)Ying, Zhang, Li, Zhou, Shao, Fei, Ma, Hong, Liu, Wang, Wang, Wu, Li, Zhou, Liu, Zhang, Zhang, Yan, Qiu, Wang, Chen, and Lin}]{DBLP:journals/corr/abs-2402-06332}
Huaiyuan Ying, Shuo Zhang, Linyang Li, Zhejian Zhou, Yunfan Shao, Zhaoye Fei, Yichuan Ma, Jiawei Hong, Kuikun Liu, Ziyi Wang, Yudong Wang, Zijian Wu, Shuaibin Li, Fengzhe Zhou, Hongwei Liu, Songyang Zhang, Wenwei Zhang, Hang Yan, Xipeng Qiu, Jiayu Wang, Kai Chen, and Dahua Lin. 2024.
\newblock \href {https://doi.org/10.48550/ARXIV.2402.06332} {Internlm-math: Open math large language models toward verifiable reasoning}.
\newblock \emph{CoRR}, abs/2402.06332.

\bibitem[{Yu et~al.(2024)Yu, Xu, Weston, and Kulikov}]{yu2024distilling21}
Ping Yu, Jing Xu, Jason Weston, and Ilia Kulikov. 2024.
\newblock \href {https://arxiv.org/abs/2407.06023} {Distilling system 2 into system 1}.
\newblock \emph{Preprint}, arXiv:2407.06023.

\bibitem[{Yuan et~al.(2024)Yuan, Pang, Cho, Li, Sukhbaatar, Xu, and Weston}]{yuan2024selfrewarding}
Weizhe Yuan, Richard~Yuanzhe Pang, Kyunghyun Cho, Xian Li, Sainbayar Sukhbaatar, Jing Xu, and Jason~E Weston. 2024.
\newblock \href {https://openreview.net/forum?id=0NphYCmgua} {Self-rewarding language models}.
\newblock In \emph{Forty-first International Conference on Machine Learning}.

\bibitem[{Zelikman et~al.(2022)Zelikman, Wu, Mu, and Goodman}]{zelikman2022star}
Eric Zelikman, Yuhuai Wu, Jesse Mu, and Noah Goodman. 2022.
\newblock \href {https://openreview.net/forum?id=_3ELRdg2sgI} {{ST}ar: Bootstrapping reasoning with reasoning}.
\newblock In \emph{Advances in Neural Information Processing Systems}.

\bibitem[{Zeng et~al.(2023)Zeng, Chen, Liu, Jiang, and Jia}]{DBLP:journals/corr/abs-2312-17080}
Zhongshen Zeng, Pengguang Chen, Shu Liu, Haiyun Jiang, and Jiaya Jia. 2023.
\newblock \href {https://doi.org/10.48550/ARXIV.2312.17080} {Mr-gsm8k: A meta-reasoning benchmark for large language model evaluation}.
\newblock \emph{CoRR}, abs/2312.17080.

\bibitem[{Zhang et~al.(2024{\natexlab{a}})Zhang, Hosseini, Bansal, Kazemi, Kumar, and Agarwal}]{zhang2024generativeverifiersrewardmodeling}
Lunjun Zhang, Arian Hosseini, Hritik Bansal, Mehran Kazemi, Aviral Kumar, and Rishabh Agarwal. 2024{\natexlab{a}}.
\newblock \href {https://arxiv.org/abs/2408.15240} {Generative verifiers: Reward modeling as next-token prediction}.
\newblock \emph{Preprint}, arXiv:2408.15240.

\bibitem[{Zhang et~al.(2024{\natexlab{b}})Zhang, Shen, Wu, Peng, Wang, Zhuang, and Lu}]{zhang-etal-2024-self-contrast}
Wenqi Zhang, Yongliang Shen, Linjuan Wu, Qiuying Peng, Jun Wang, Yueting Zhuang, and Weiming Lu. 2024{\natexlab{b}}.
\newblock \href {https://doi.org/10.18653/v1/2024.acl-long.197} {Self-contrast: Better reflection through inconsistent solving perspectives}.
\newblock In \emph{Proceedings of the 62nd Annual Meeting of the Association for Computational Linguistics (Volume 1: Long Papers)}, pages 3602--3622, Bangkok, Thailand. Association for Computational Linguistics.

\bibitem[{Zheng et~al.(2023)Zheng, Chiang, Sheng, Zhuang, Wu, Zhuang, Lin, Li, Li, Xing, Zhang, Gonzalez, and Stoica}]{NEURIPS2023_91f18a12}
Lianmin Zheng, Wei-Lin Chiang, Ying Sheng, Siyuan Zhuang, Zhanghao Wu, Yonghao Zhuang, Zi~Lin, Zhuohan Li, Dacheng Li, Eric Xing, Hao Zhang, Joseph~E Gonzalez, and Ion Stoica. 2023.
\newblock \href {https://proceedings.neurips.cc/paper_files/paper/2023/file/91f18a1287b398d378ef22505bf41832-Paper-Datasets_and_Benchmarks.pdf} {Judging llm-as-a-judge with mt-bench and chatbot arena}.
\newblock In \emph{Advances in Neural Information Processing Systems}, volume~36, pages 46595--46623. Curran Associates, Inc.

\bibitem[{Zhong et~al.(2023)Zhong, Cui, Guo, Liang, Lu, Wang, Saied, Chen, and Duan}]{zhong2023agievalhumancentricbenchmarkevaluating}
Wanjun Zhong, Ruixiang Cui, Yiduo Guo, Yaobo Liang, Shuai Lu, Yanlin Wang, Amin Saied, Weizhu Chen, and Nan Duan. 2023.
\newblock \href {https://arxiv.org/abs/2304.06364} {Agieval: A human-centric benchmark for evaluating foundation models}.
\newblock \emph{Preprint}, arXiv:2304.06364.

\end{thebibliography}

\newpage
\appendix
\setcounter{table}{0}   
\setcounter{figure}{0}
\renewcommand{\thetable}{A\arabic{table}}
\renewcommand{\thefigure}{A\arabic{figure}}

\section{Appendix}

\subsection{Training \& Evaluation Details}
\label{appendix:more_details}
\subsubsection{Critic Data Construction}
\label{appendix:Critic Data Construction}
\paragraph{GSM8K} On GSM8K, since \texttt{GPT-4} already got 92.0\% accuracy on the test set \citep{DBLP:journals/corr/abs-2303-08774}, which makes it hard to obtain negative data, we use \texttt{GPT-3.5-Turbo-0125} instead to sample 10 solutions for each question in the training set. Then, we use \texttt{GPT-4-Turbo} as the critic-refine model to criticize the solutions (Table \ref{table:prompt_gsm8k}), with $K=16$ retry. We obtain 63,485 cases, with 49,832 positive examples and 13,653 negative examples.

In the second stage of GSM8K critique construction, we use the learned critic model to repeatedly sample until we obtain at most 5 positive and 5 negative solutions. For strong LLMs like \texttt{LLaMA-3}, it's challenging to get enough negative solutions even among 512 samples, so the size of negative data would be slightly smaller. Then, we use the learned critic model to criticize itself, also with $K=16$ retry. In stage two, we obtain 62,877 instances, with 39,654 positive and 26,001 negative. Among the two stages, we got 126,362 instances, with 86,708 positive and 39,654 negative.

\paragraph{MATH} On MATH, in the first stage, we directly use the 90,074 \texttt{GPT-4} generated solutions of PRM800K Dataset \citep{lightman2024lets}, with 11,665 positive instances which all the step labels are correct, and 78,409 negative instances which one step label is incorrect. Since the MATH dataset is challenging, in order to reduce retry of \texttt{GPT-4-Turbo} and avoid not getting valid critique, for the critique of the negative solution,  we additionally append reference solution in the input prompt, and hint it might contain mistakes, as suggested in prior work \citep{zelikman2022star}; 
for the positive solution, we simply hint it's correct. 
After obtaining the initial critique, we use \texttt{GPT-4-Turbo} again to remove hint phrases like ``According to the reference'' or ``Given the hint'' since we do not have any hint or reference during the test time. In stage one, we obtain 1,606 positive cases and 69,775 negative cases.

Similarly, in the second stage of MATH, we use the learned critic model to sample at most 5 positive and negative solutions. Then, we first use the critic model itself to critic its solutions, and without any hints, under $K=16$ retry, and use \texttt{GPT-4-Turbo} to retry another $K=16$ times with hint if failed. We construct 51,618 positive cases and 65,456 negative cases. Among the two stages, we got 188,455 cases, with 53,224 positive and 135,231 negative.

\subsubsection{Answer Extraction}
\label{appendix:Answer Extraction}

We let the model print the answer in the format \texttt{\textbackslash boxed\{answer\}}. The model generates the answer following this pattern. We then extract the regular expression \texttt{\textbackslash boxed\{.*\}} from the model output, and obtain the valid answer expression with matched parenthesis. The Python code for answer extraction is shown in Table \ref{tab:python_code}.

\subsubsection{Evaluation Metric}
\label{appendix:metric}
\paragraph{Solution} 
For the evaluation of the solution, we extract the final answer (Appendix \ref{appendix:Answer Extraction}) and compute the metrics of Top-1 Accuracy \texttt{Acc} and Refine Accuracy \texttt{Refine-Acc}, in which the original Top-1 predict-answer is replaced with a refined one if the critic model found an error and made iterative refinement (Section \ref{sec:Iterative Refine}). We also compute Majority Vote Accuracy \texttt{Maj1@N} \citep{wang2023selfconsistency} and Majority Vote Accuracy After Critique \texttt{Critic + Maj1@N} (Section \ref{sec:Critic As Filter}), which is to select the most frequent answer among $N$ samples, i.e. $\arg \max _a \sum_{i=1}^N \mathds{1}\left(\mathbf{a}_i=a\right)$. Following \citet{DBLP:journals/corr/abs-2310-10047, havrilla2024teaching}, we compute \texttt{Pass@N}, which select the gold answer $g$ among the $N$ predictions if present, i.e. $\arg \max _a  \mathds{1}\left(g=a\right)$.

\paragraph{Critique}
For the ``evaluation of evaluation'', we compute Precision, Recall, and F1 for error detection; also, we compute Critic Accuracy, where the critique should find the error in wrong answer solutions and pass the correct answer solution:
\begin{equation}
P = \frac{ \{| Pred_i \neq Ans_i \land -1 \in L_i |\} }{ |\{-1 \in L_i \}| } \end{equation}

\begin{equation}
R = \frac{ \{| Pred_i \neq Ans_i \land -1 \in L_i |\} }{ |\{Pred_i \neq Ans_i\}| }
\end{equation}

\begin{equation}
F1 = \frac{2 * P * R}{P + R} 
\end{equation}

\begin{equation}
\begin{split}
& CriticAcc = \\
& \frac{1}{N} \sum_{i=1}^N (Pred_i = Ans_i \land -1 \notin L_i ) \\
& \lor ( Pred_i \neq Ans_i \land -1 \in L_i )  
\end{split}
\end{equation}
Here, for the $i$-th instance, $Pred_i$ is the prediction answer, $Ans_i$ is the ground truth answer, and $L_i$ is the predicted step label list.

\subsubsection{Implementation Details}
\label{appendix:Training Details}
The training on GSM8K dataset takes 23 hours, while on MATH dataset it takes 37 hours. The two datasets are under MIT license. For software, Huggingface Transformers, DeepSpeed, vLLM, and FastChat libraries we used are under Apache-2.0 license.

\subsection{Additional Result on MATH}
Table \ref{table:Solution Accuracy of MATH} presents the results of Critic-CoT training on MATH dataset, with the original 7,500/5,000 split setting.

\begin{table*}[h]
\small
\centering
\begin{tabularx}{0.82\textwidth}{lll}
\toprule
\textbf{Model} & \textbf{Sampling Method} & \textbf{Acc.} \\
\midrule
Llama-3-70B-Instruct \citep{llama3modelcard} & - & 51.0 \\
& Maj1@96 & 63.5  \\
& Maj1@512 & 64.3  \\
Llama-3.1-70B-Instruct \citep{llama3modelcard} & - & 68.0 \\
DeepSeek-V2 Chat-236B \citep{DBLP:journals/corr/abs-2405-04434} & - & 53.9 \\
Qwen2-72B \citep{yang2024qwen2technicalreport} & - & 69.0 \\
GPT4-0314 \citep{DBLP:journals/corr/abs-2303-08774} & - & 42.5 \\
\midrule
\textbf{Critic-CoT, Llama-3-70B-Instruct (\textit{Ours})} & - & \textbf{56.2} \\
 & Iterative Refine & \textbf{56.6} {\color[HTML]{32cb00}$\uparrow$ 0.4}  \\
 & Maj1@96 & 64.2   \\
 & Critic + Maj1@96 & \textbf{65.0} {\color[HTML]{32cb00}$\uparrow$ 0.8}  \\
 & Maj1@512 & 64.4  \\
 & Critic + Maj1@512 & \textbf{66.4} {\color[HTML]{32cb00}$\uparrow$ 2.0} \\

\bottomrule
\end{tabularx}
\caption{Solution Accuracy of MATH. The top-1 accuracy of our method increases from 51.0\% to 56.2\%, and the effect of iterative refinement is moderate but positive improvement of 0.4\%, while the performance gain of the critic filter is larger.}
\label{table:Solution Accuracy of MATH}

\end{table*}

\subsection{Analysis}
\label{appendix:analysis}

\subsubsection{Critic Performance}

For both datasets, the critic model's accuracy continues to grow as the sample size $N$ increases, ultimately surpassing the performance of the majority vote, which gradually converges. Specifically, in the MATH dataset, the critic model achieves substantially higher accuracy than the solution accuracy, consistently outperforming the naive majority vote due to the critic filter's superior performance. This stark contrast highlights the critic model's effectiveness in identifying and promoting correct answers. In the GSM8K dataset, despite having a critic accuracy of only 92.3\%, the critic model still manages to deliver higher accuracy gains. This outcome suggests that the critic model successfully filters answers to increase the density of correct answers and decrease the density of wrong answers, compared to the normal answer distribution. The overall results demonstrate the critic model's robust capability to enhance accuracy across different datasets, validating its practical utility in improving prediction outcomes.

\begin{figure*}[htbp]
    \centering
    \includegraphics[width=1.0\textwidth]{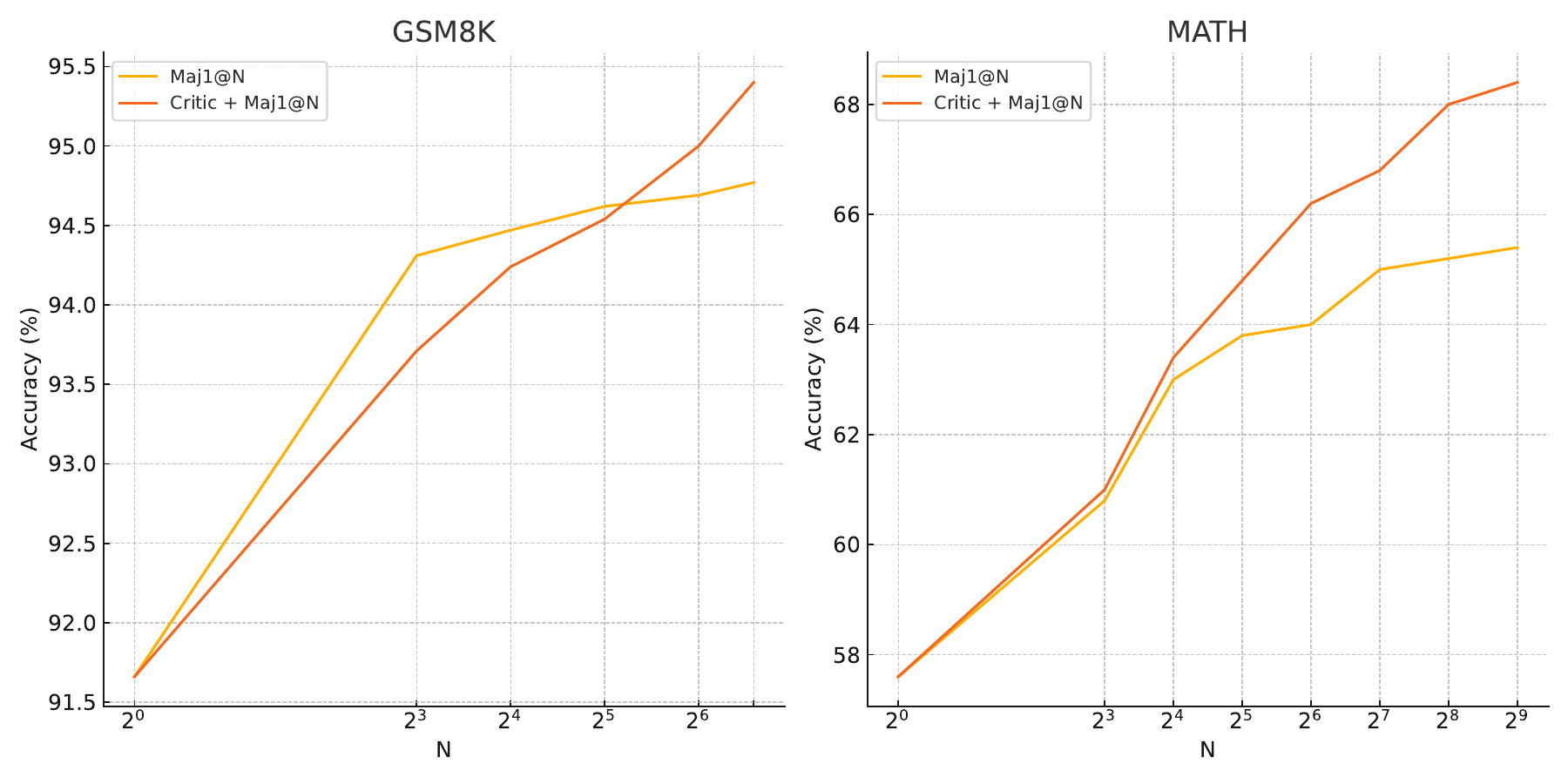}
    \caption{Performance of majority vote on GSM8K and MATH500 Datasets}
    \label{fig:gsm8k_math_plots}
\end{figure*}

\subsubsection{Inspect on Iterative Refine}

\begin{table}[h]

    \begin{subtable}[h]{0.45\textwidth}
        \centering
        \small
        \begin{tabular}{@{}cccc@{}}
        \toprule
        \textbf{Round} & \textbf{Refine Acc.} & \textbf{\begin{tabular}[c]{@{}l@{}}True $\rightarrow$ \\ True \end{tabular}} & \textbf{\begin{tabular}[c]{@{}l@{}}False $\rightarrow$ \\ True \end{tabular}} \\ \midrule
        0 & 91.7 & - & - \\
        1 & 91.7 & 48.2 & 45.3 \\
        2 & 92.6 & 78.6 & 37.5 \\
        3 & 92.7 & 64.3 & \textbf{53.1} \\
        4 & 93.0 & 73.2 & 50.0 \\
        5 & 93.2 & 75.0 & \textbf{53.1} \\
        6 & 93.2 & 76.8 & \textbf{53.1} \\
        7 & \textbf{93.3} & \textbf{80.4} & 50.0 \\
        8 & \textbf{93.3} & \textbf{80.4} & 50.0 \\ \bottomrule
        \end{tabular}
        \caption{GSM8K}
        \label{subtable:Iterative Refine on GSM8K}
    \end{subtable}
    \hfill
    \begin{subtable}[h]{0.45\textwidth}
        \centering
        \small
        \begin{tabular}{@{}cccc@{}}
        \toprule
        \textbf{Round} & \textbf{Refine Acc.} & \textbf{\begin{tabular}[c]{@{}l@{}}True $\rightarrow$ \\ True \end{tabular}} & \textbf{\begin{tabular}[c]{@{}l@{}}False $\rightarrow$ \\ True \end{tabular}} \\ \midrule
        0 & 57.6 & - & - \\
        1 & 53.4 & 29.0 & 17.7 \\
        2 & 57.2 & \textbf{65.7} & 13.9 \\
        3 & 55.2 & 48.6 & \textbf{15.2} \\
        4 & 57.2 & 60.9 & 15.9 \\
        5 & 57.4 & 60.0 & 17.1 \\
        6 & 57.6 & 61.4 & 17.1 \\
        7 & \textbf{57.8} & 60.0 & \textbf{18.4} \\
        8 & \textbf{57.8} & 62.9 & 16.5 \\ \bottomrule
        \end{tabular}        \caption{MATH500}
        \label{subtable:Iterative Refine on MATH}
     \end{subtable}
     \caption{Iterative Refine on GSM8K (\ref{subtable:Iterative Refine on GSM8K}) and MATH500 (\ref{subtable:Iterative Refine on MATH}). }
     \label{tab:Iterative Refine}
\end{table}

The iterative refinement process for the GSM8K and MATH datasets demonstrates different levels of effectiveness due to their complexity, as shown in Table \ref{tab:Iterative Refine}. GSM8K, being simpler, shows a higher success rate in refinement. For effective refinement, the number of false answers corrected (False \(\rightarrow\) True) must exceed the number of true answers incorrectly changed (True \(\rightarrow\) False). Despite occasional mistakes by the critic, correct answers are not always altered incorrectly.

For GSM8K (Table \ref{subtable:Iterative Refine on GSM8K}), accuracy improves from 91.7\% initially to 93.3\% by the seventh round, with significant gains in both true-to-true and false-to-true transformations. In contrast, MATH (Table \ref{subtable:Iterative Refine on MATH}) starts at 57.6\% accuracy, reaching 57.8\% by the seventh round. The iterative refinement process tends to converge, which is expected.

\begin{figure*}[h]
    \centering
    \includegraphics[width=\textwidth]{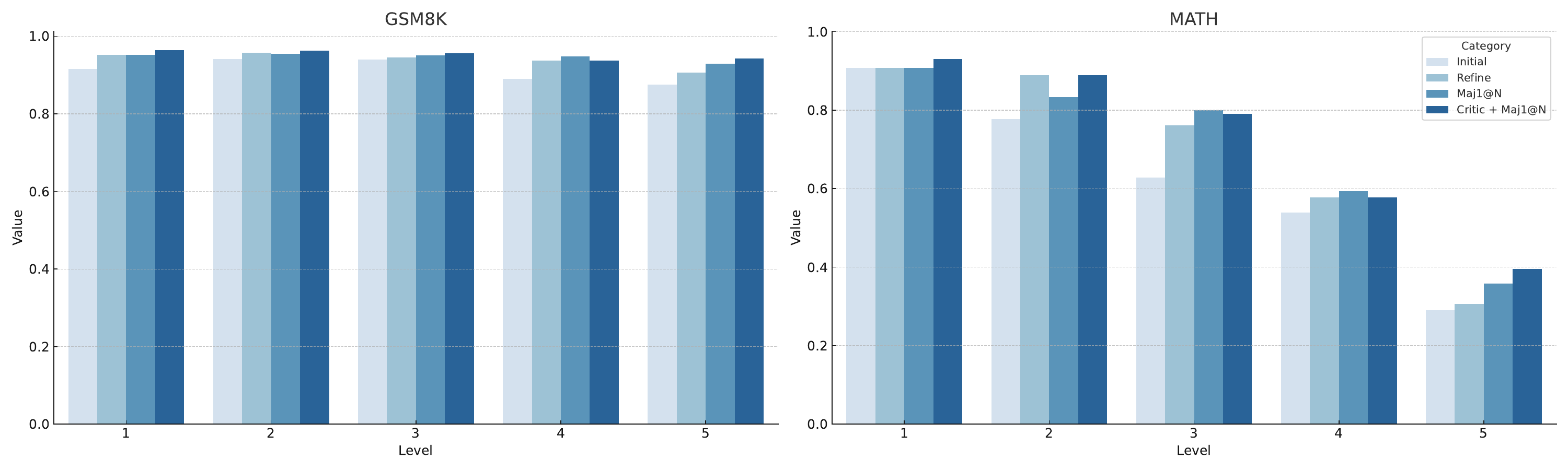}
    \caption{Performance group by difficulty level, on GSM8K and MATH500 Datasets}
    \label{fig:gsm8k_math_level_plots}
\end{figure*}

\begin{table*}[h]
\begin{tabularx}{\textwidth}{@{}llll|llll@{}}
\toprule
\multicolumn{4}{c|}{\textbf{GSM8K}} & \multicolumn{4}{c}{\textbf{MATH}} \\ 
\textbf{Round} & \textbf{Acc.} & \textbf{Avg. Token} & \textbf{Avg. Latency (s)} & \textbf{Round} & \textbf{Acc.} & \textbf{Avg. Token} & \textbf{Avg. Latency (s)} \\ \midrule
0 & 91.7 & 176.7 & 1.156 & 0 & 57.6 & 296.7 & 1.941 \\
1 & 91.7 & 377.6 & 2.470 & 1 & 53.4 & 694.5 & 4.544 \\
2 & 92.6 & 395.1 & 2.585 & 2 & 57.2 & 872.8 & 5.710 \\
3 & 92.7 & 408.2 & 2.671 & 3 & 55.2 & 1003.0 & 6.562 \\
4 & 93.0 & 418.5 & 2.738 & 4 & 57.2 & 1114.4 & 7.291 \\
5 & 93.2 & 426.3 & 2.789 & 5 & 57.4 & 1202.0 & 7.864 \\
6 & 93.2 & 432.6 & 2.830 & 6 & 57.6 & 1282.7 & 8.392 \\
7 & 93.3 & 436.8 & 2.858 & 7 & 57.8 & 1343.1 & 8.787 \\
8 & 93.3 & 440.4 & 2.881 & 8 & 57.8 & 1390.6 & 9.098 \\ \bottomrule
\end{tabularx}
     \caption{Inference cost of Iterative Refine on GSM8K and MATH500. }
     \label{tab:Iterative Refine Inference Cost}

\end{table*}

\begin{table*}[h]
\begin{tabularx}{\textwidth}{@{}llll|llll@{}}
\toprule
\multicolumn{4}{c|}{\textbf{GSM8K}} & \multicolumn{4}{c}{\textbf{MATH}} \\ 
\textbf{Sample} & \textbf{Acc.} & \textbf{Avg. Token} & \textbf{Avg. Latency (s)} & \textbf{Sample} & \textbf{Acc.} & \textbf{Avg. Token} & \textbf{Avg. Latency (s)} \\ \midrule
1 & 91.7 & 176.7 & 1.156 & 1 & 57.6 & 296.7 & 1.941 \\
8 & 93.7 & 2974.6 & 2.433 & 8 & 61.0 & 4446.33 & 3.636 \\
16 & 94.2 & 6261.6 & 5.121 & 16 & 63.4 & 9198.43 & 7.523 \\
32 & 94.5 & 12524.1 & 10.242 & 32 & 64.8 & 19228.69 & 15.726 \\
64 & 95.0 & 25057.9 & 20.493 & 64 & 66.2 & 41709.00 & 34.110 \\
96 & 95.4 & 37574.6 & 30.729 & 96 & 66.6 & 60907.41 & 49.811  \\ \bottomrule
\end{tabularx}
     \caption{Inference cost of Critic As Filter on GSM8K and MATH500. }
     \label{tab:Critic As Filter Inference Cost}
\end{table*}

\subsubsection{Group By Difficulty Level}

For the MATH dataset, the difficulty level is given from 1 to 5. For the GSM8K dataset, we set the difficulty level according to the number of expressions $n$ that appeared in the reference solution, i.e., $max(1, min(5, n))$. As illustrated in Figure \ref{fig:gsm8k_math_level_plots}, the performance on the GSM8K dataset shows a gradual decline as the difficulty level increases. This trend is accompanied by the emerging effects of the critic and refine stages, which become more prominent at higher difficulty levels. In contrast, the accuracy on the MATH dataset declines sharply as the problems become more challenging. Generally, the refine stage proves effective across all levels, while the critic stage is beneficial at most levels, with some minor exceptions. These observations suggest potential areas for further improvements in the critic mechanism.
\subsection{Discussion}
\label{appendix:discussion}

\subsubsection{Discriminative Verifier for Mathematics}

To further improve the reasoning ability of large language models, one applicable approach is through the use of reward models, which can either be used in reinforcement learning during training \citep{NEURIPS2022_b1efde53} or rejection sampling at test time \citep{cobbe2021trainingverifierssolvemath}. While outcome-supervised reward models (ORMs) allow for the automatic collection of training data based on the signal of the gold answer, process-supervised reward models (PRMs) would be more advantageous for more precise feedback, better interpretability and stronger alignment \citep{lightman2024lets}. 

To reduce the considerable human labeling cost and difficulty for dense annotation, a series of works based on automatic approaches have been proposed \citep{DBLP:journals/corr/abs-2312-08935, DBLP:journals/corr/abs-2405-03553, luo2024improvemathematicalreasoninglanguage, snell2024scalingllmtesttimecompute}, all under the heuristic that for an incorrect solution, the first error step is where the continuation of previous step would lead to a correct answer. This may bring noise into training data due to false positives and negatives \citep{luo2024improvemathematicalreasoninglanguage}. Moreover, annotation based on the implicit solution continuation alone does not leverage LLM's emerging ability of critic, which is in a more explicit and analytic way and brings better explainability \citep{saunders2022selfcritiquingmodelsassistinghuman, yuan2024selfrewarding, luo2023critiqueabilitylargelanguage, mcaleese2024llmcriticshelpcatch}. Additionally, binary 0/1 discrimination alone, whether outcome-based or process-based, remains more similar to System-1 reasoning rather than the desirable System-2, thus may not fully leverage the computation power support by empirically successful Chain-of-Thought prompting \citep{feng2023towards, li2024chain}.

\subsubsection{Differences Between Critic-CoT and Reflexion}
We adopt a similar approach to Relexion \citep{shinn2023reflexion}, which leverages natural language critique to facilitate refinement, but our method diverges in the following ways:

\paragraph{Step-wise CoT Critique} Reflexion translates and augments the binary reward signal from the environment to natural language, but on an instance level. Instead, fine-grained Chain-of-Thought analysis at the step level, which is more systematic, and enables us to locate the error and start refinement from a specific step, rather than refine the whole attempt.

\paragraph{Enhanced Critic ability} While Relexion proposed an in-context learning pipeline for policy optimization under the oracle success/fail binary feedback signal, \citet{huang2024large} showed that without external feedback, vanilla LLMs cannot self-correct effectively due to limited critique ability. Therefore, to teach the LLMs the ability of intrinsic self-critique, our approach tries to learn the critique ability itself, through Critic-CoT training, and can apply it to test-time situations where the oracle feedback signal is not available.




\subsubsection{Differences between Critic-CoT and Long Reasoning Model}
Compared with these reasoning models, our method differs in terms of training mechanism and inference method. While reasoning models attempt to improve the accuracy with longer CoT output, our Critic-CoT method introduces explicit, step-wise critique, which helps identify the error more precisely and facilitates refinement. The key distinctions are as follows:
  \paragraph{Controllability and Efficiency} It could be non-trivial to control the long reasoning models to critic and refine during inference. When to critic, critic at which step, and when to restart depends on the model itself, and they often exhibit "overthinking," performing unnecessary reflections or exploring multiple solutions even for simple problems, leading to increased computational costs without proportionate gains \citep{chen2024not}. On the contrary, our method can achieve controllable test-time scaling, by specifying the number of reflection rounds for Iterative Refine, and the number of samples for Critic As Filter. 
  \paragraph{Adaptability} While models like R1 \citep{guo2025deepseek} are trained with reinforcement learning with outcome reward and implicitly learn the critic and refine patterns, this self-reflection behavior is not guaranteed to emerge reliably across all models \citep{gandhi2025cognitive}. Instead, our proposed framework explicitly teaches the model step-level critique and refinement, and can be easily adapted to different LLMs of various sizes.

\textbf{Moreover, we believe our approach is complementary to long reasoning models.} Given the limited self-critic ability of current R1-like models \citep{he2025can}, our method could strengthen these models to precisely detect errors while maintaining confidence in correct steps. Reasoning models with enhanced self-critic ability could potentially reduce overthinking, boost performance for more complex tasks \citep{phan2025humanity}, and move toward self-improvement and scalable oversight \citep{schaul2024boundless}.

\subsubsection{Comparison Between Inference Methods}
\label{Appendix:Comparison Between Inference Methods}

The statistics of average tokens and latency per instance are shown in Table \ref{tab:Iterative Refine Inference Cost} and \ref{tab:Iterative Refine Inference Cost}.
For Iterative Refinement, as the number of rounds increases, the computation cost and latency slowly increase, but the performance gain gradually becomes saturated. The reason for non-linear computation increase is that, refinement only occurs if the critique detects that the attempt is wrong and stops if the critique validates it. Yet, the pipeline of iterative refinement could be more sensitive to error accumulation, which could limit performance.

For Critic As Filter, the computation cost and latency increase linearly as the number of samples increases, and the performance also improves steadily. The linear cost increase is because we need to sample the attempt first, then perform CoT critic for each step. To reduce latency, we can parallelize the inference, and in our setting, we use 32 GPUs for inference, which deploys 8 models in total.

As the results in Table \ref{table:Solution Accuracy of GSM8K and MATH500} shows, the performance of majority vote and Critic As Filter surpass Iterative Refinement. We believe it's due to the intrinsic challenges of refinement and the relatively limited search space.

On the one hand, for Iterative Refinement to work properly, it requires the model to 1) Detect errors on an attempt; 2) Refine the mistakes; 3) Exit if no further errors are detected. This pipeline could be more sensitive to error accumulation. Moreover, it only edits on a single example and has a limited retry, which is sample-efficient, but may not explore the solution space more actively, as majority vote does. Specifically on the dataset of GSM8K, the invocation statistics are as follows:
\begin{itemize}
    \item Majority vote: 1319 * 96 = 126,624
    \item Iterative Refinement: among 1319 test cases, our Critic-CoT model predicts 274 problematic instances and iterates 1627 times (on average 5.94 rounds for each wrong case), which makes in total 1319 * 2 + 1627 * 2 = 5892 invocations, which is 21.5 times fewer than Majority Vote calls.

\end{itemize}
On the other hand, majority vote is a strong baseline, as it requires massive sampling. It leverages diverse reasoning paths and tries to mitigate the stochastic of a single sample. But under the method of Critic As Filter, we actively filter out problematic attempts and perform the majority vote on the more reasonable candidates, rather than equally account for all the predicted answers as the vanilla majority vote does, which further increases performance. This in turn demonstrates our model's strong ability to critique.



\subsubsection{Self-Reflection}
Besides the main results, through out-of-domain evaluation in Table \ref{tab:OOD solution}, we find our model demonstrates generalized ability to critique and refine.
While the ability of LLMs to self-reflect still remains an open question, and we hope our work as a valuable exploration could shed light on future studies in this area.

Moreover, as long as we adequately improve the models' ability to critique, we could achieve test-time performance increase in the form of "self-reflection". As the experiment results present, after Critic-CoT training, the ability to critique and generate both improves, though they are not exactly identical. Notably, the critique ability can surpass the task-solving ability, allowing the model to detect errors even when it has a low probability of generating a valid solution, as prior works \citep{saunders2022selfcritiquingmodelsassistinghuman, lin-etal-2024-criticbench} also suggest. This indicates that by strengthening the model’s CoT critique ability beyond its generation capability, we can leverage this discriminative power to reject imperfect responses and achieve positive performance gains.


\subsubsection{Process Correctness of Correct Answer Attempt}
\label{appendix:Process Correctness of Correct Answer Attempt}

We each sample 100 correct answer solutions, on GSM8K by GPT-3.5-Turbo and GPT-4-Turbo, and MATH by GPT-4-Turbo, and manually check if all intermediate steps are correct. 
The results are demonstrated in Table \ref{table:Human Evaluation on the Intermediate Steps of Attempts with Correct Answer}. We find that in general, the correct final answer is a good indicator of correct intermediate steps. Also, from GSM8K to MATH, as the reasoning traces become longer and more complicated, the percentage of correct answer but with wrong intermediate steps increases.

\subsubsection{Alternatives of Leveraging Proprietary Models for Data Construction }

To mitigate the data bias from GPT-4, we can introduce other open-source LLMs like DeepSeek-V3 or Qwen2.5-Max to perform multi-agent debates \citep{chan2024chateval, liang-etal-2024-encouraging}. This could encourage diverse thinking trajectories and overcome the inherited bias from a single model. 

To reduce computation costs, we may substitute with smaller models. While our weak supervision method can effectively filter out problematic critic, to maintain a high success rate of generating valid data for smaller models, we can apply iterative refinement with more rounds, Monte Carlo Tree Search that more efficiently explores the search space, and other techniques. This can facilitate smaller models to produce desirable critic-refine data that is comparable with larger models.

\begin{table*}[ht]
    \centering
    \begin{tabular}{|l|}
        \hline
        \texttt{import re} \\
        \\
        \texttt{def extract\_boxed\_expressions\_custom(text):} \\
        \quad \texttt{stack = []} \\
        \quad \texttt{current\_expr = ""} \\
        \quad \texttt{i = 0} \\
        \quad \texttt{while i < len(text):} \\
        \qquad \texttt{if text[i:i+7] == r"\textbackslash boxed\{":} \\
        \qquad\qquad \texttt{if stack:} \\
        \qquad\qquad\qquad \texttt{current\_expr += text[i]} \\
        \qquad\qquad \texttt{stack.append("\{")} \\
        \qquad\qquad \texttt{i += 7} \\
        \qquad \texttt{elif text[i] == "\{" and stack:} \\
        \qquad\qquad \texttt{stack.append("\{")} \\
        \qquad\qquad \texttt{current\_expr += text[i]} \\
        \qquad\qquad \texttt{i += 1} \\
        \qquad \texttt{elif text[i] == "\}" and stack:} \\
        \qquad\qquad \texttt{stack.pop()} \\
        \qquad\qquad \texttt{if stack:} \\
        \qquad\qquad\qquad \texttt{current\_expr += text[i]} \\
        \qquad\qquad \texttt{else:} \\
        \qquad\qquad\qquad \texttt{current\_expr = re.split('=', current\_expr)[-1]} \\
        \qquad\qquad\qquad \texttt{return current\_expr} \\
        \qquad\qquad \texttt{i += 1} \\
        \qquad \texttt{elif stack:} \\
        \qquad\qquad \texttt{current\_expr += text[i]} \\
        \qquad\qquad \texttt{i += 1} \\
        \qquad \texttt{else:} \\
        \qquad\qquad \texttt{i += 1} \\
        \quad \texttt{return ''} \\
        \\
        \texttt{def extract\_answer(text):} \\
        \quad \texttt{predict\_answer\_span = re.findall(r'(\textbackslash boxed\{.*\})', text, re.DOTALL)[-1]} \\
        \quad \texttt{predict\_answer = extract\_boxed\_expressions\_custom(predict\_answer\_span)} \\
        \quad \texttt{return predict\_answer} \\
        \hline
    \end{tabular}
    \caption{Python code for extracting the boxed answer.}
    \label{tab:python_code}
\end{table*}

\begin{table*}[h]
\centering
\begin{tabularx}{0.95\textwidth}{@{}lll@{}}
\toprule
\textbf{Data} & \textbf{Model} & \textbf{Intermediate Accuracy of Correct Answer Attempt} \\ \midrule
GSM8K & GPT-3.5-Turbo & 97\% \\
GSM8K & GPT-4-Turbo & 99\% \\
MATH & GPT-4-Turbo & 93\% \\ \bottomrule
\end{tabularx}
\caption{Human Evaluation on the Intermediate Steps of Attempts with Correct Answer}
\label{table:Human Evaluation on the Intermediate Steps of Attempts with Correct Answer}

\end{table*}






\subsection{Examples of Refinement}
\label{appendix:An Example of Refinement on GSM8K}
As presented in Figure \ref{fig:example_GSM8K}, in an example of GSM8K, the model forgot to add one year at Step 3; then, through CoT critique, the model found that while Step 1 and Step 2 are correct, Step 3 contains this ignorance error. Finally, guided by the critique of Step 3, the model made a correction and reached the gold answer of 13. Similarly, in an example of MATH (\ref{fig:example_MATH}), the model identified and successfully fixed the error.

\subsection{Prompts}
\label{appendix:Prompts}
Table \ref{table:prompt_gsm8k}, Table \ref{table:prompt_math_correct}, and Table \ref{table:prompt_math_incorrect} presents the prompt for critic-refine data collection using GPT4-Turbo, with Table \ref{table:prompt_math_incorrect_remove_hint} for removing the hint phrases (Section \ref{section:Two-Stage Training}). Table \ref{table:prompt_solve}, Table \ref{table:prompt_critic}, and Table \ref{table:prompt_refine} shows the prompt of trained model for solving, critique, and refinement during stage-2-training (Section \ref{section:Two-Stage Training}) and inference (Section \ref{section:Inference}). Table \ref{table:Full example of Critic-CoT}, Table \ref{table:Full example of Step-wise Label Critic} and Table \ref{table:Full example of Final Label Critic} present the prompts and responses of a single-turn critique-refinement, under Critic-CoT, Step-wise Label Critic and Final Label Critic respectively. Table \ref{table:Full example of Critic-CoT on HumanEval} presents a single-turn critique-refinement on HumanEval.


\onecolumn

\begin{figure}[h]
\begin{tcolorbox}[colback=blue!5!white,colframe=blue!75!black]
\begin{small}
\textbf{\texttt{Question:}}
\\
Carlos is planting a lemon tree. The tree will cost \$90 to plant. Each year it will grow 7 lemons, which he can sell for \$1.5 each. It costs \$3 a year to water and feed the tree. How many years will it take before he starts earning money on the lemon tree?\vspace{3mm}

\textbf{\texttt{Model Attempt:}}
\\
Step 1:
Calculate the annual income from selling lemons. This is done by multiplying the number of lemons grown per year (7) by the price per lemon (\$1.5). So, 7 * \$1.5 = \$10.50.

Step 2:
Subtract the annual cost of watering and feeding the tree (\$3) from the annual income from selling lemons (\$10.50). So, \$10.50 - \$3 = \$7.50. This is the net income per year after expenses.

Step 3:
Determine when Carlos will start earning money back. 
This is done by dividing the initial cost of planting the tree (\$90) by the net income per year (\$7.50). So, \$90 / \$7.50 = 12 years. 

Therefore, it will take Carlos \boxed{ {\color[HTML]{FF0000} 12} } years before he starts earning money on the lemon tree.

\end{small}
\DrawLine
\vspace{3mm}

\begin{small}
\textbf{\texttt{Model Critique:}}
\\
Step 1: The attempt calculates the annual income from selling lemons by multiplying the number of lemons grown per year (7) by the price per lemon (\$1.5), resulting in \$10.50. 
\\
Conclusion: Step 1 is correct.
\\
\\
Step 2: The attempt subtracts the annual cost of watering and feeding the tree (\$3) from the annual income from selling lemons (\$10.50), resulting in a net income per year of \$7.50. 
\\
Conclusion: Step 2 is correct.
\\
\\
Step 3: The attempt divides the initial cost of planting the tree (\$90) by the net income per year (\$7.50), resulting in 12 years. However, this calculation does not account for the fact that Carlos will start earning money back in the first year, even though he won't break even until the 13th year. \\
\textbf{\textit{Conclusion: Step 3 is incorrect.}}

\end{small}

\DrawLine
\vspace{3mm}

\begin{small}
\textbf{\texttt{Model Refinement:}}
\\
{\color[HTML]{32cb00}
Step 3 (Corrected): Determine when Carlos will start earning money back. This is done by dividing the initial cost of planting the tree (\$90) by the net income per year (\$7.50). So, \$90 / \$7.50 = 12 years. 
\\
However, since Carlos will start earning money back in the first year, we need to add one more year to this calculation. Therefore, it will take Carlos \boxed{13} years before he starts earning money on the lemon tree.
}

\end{small}
\end{tcolorbox}
\caption{\small
Criticize and refine a problem in the GSM8K test set.}
\label{fig:example_GSM8K}
\end{figure}

\begin{figure}[t]
\begin{tcolorbox}[colback=blue!5!white,colframe=blue!75!black]
\begin{small}

\textbf{\texttt{Question:}}
\\
What is the smallest positive perfect cube that can be written as the sum of three consecutive integers?
\vspace{3mm}

\textbf{\texttt{Model Attempt:}}
\\
Step 1:
Let the three consecutive integers be \(x-1\), \(x\), and \(x+1\). The sum of these three integers is \(3x\).

Step 2:
Since we are looking for a perfect cube, let's express \(3x\) as a cube of some integer \(n\), where \(n^3 = 3x\).

Step 3:
To find the smallest such \(n\), we need to minimize \(x\). 
{
\color[HTML]{FF0000}
However, since \(n^3\) must be divisible by 3, and \(x\) is an integer, \(n\) itself must also be divisible by 3.
}

Step 4:
The smallest integer \(n\) that is divisible by 3 and gives us a perfect cube when cubed is \(n=6\), because \(6^3 = 216\).

Step 5:
Substituting \(n=6\) into \(n^3 = 3x\), we get \(216 = 3x\), which simplifies to \(x = 72\).
\\
Step 6:
Therefore, the smallest positive perfect cube that can be written as the sum of three consecutive integers is 
\(\boxed{ {\color[HTML]{FF0000} 216} }\).
\\
\end{small}
\DrawLine
\vspace{3mm}

\begin{small}
\textbf{\texttt{Model Critique:}}
\\
Step 1: Let the three consecutive integers be \(x-1\), \(x\), and \(x+1\). The sum of these three integers is indeed \(3x\).
\\
Conclusion: Step 1 is correct.
\\
\\
Step 2: Expressing \(3x\) as a cube of some integer \(n\), where \(n^3 = 3x\), correctly sets up the equation needed to solve the problem. This step accurately translates the problem statement into a mathematical expression.
\\
Conclusion: Step 2 is correct.
\\
\\
Step 3: The claim that \(n\) must be divisible by 3 because \(n^3\) must be divisible by 3 is a misunderstanding. In reality, \(n^3\) being divisible by 3 does not imply that \(n\) itself must be divisible by 3; rather, it simply means that \(n^3\) will always be divisible by 3 regardless of whether \(n\) is divisible by 3. This step introduces an unnecessary restriction on \(n\).
\\
\textbf{\textit{Conclusion: Step 3 is incorrect.}}

\end{small}
\DrawLine
\vspace{3mm}

\begin{small}
\textbf{\texttt{Model Refinement:}}
\\
{\color[HTML]{32cb00}
Step 3 (Corrected): Since we are looking for the smallest positive perfect cube that can be expressed as \(3x\), we should start by identifying the smallest perfect cubes and checking if any of them can be represented in this form. 
}

Step 4: The smallest perfect cubes are \(1^3=1\), \(2^3=8\), \(3^3=27\), \(4^3=64\), \(5^3=125\), \(6^3=216\), and so on. Among these, \(27\) is the first number that can be expressed as \(3x\), where \(x\) is an integer (\(x=9\)).

Step 5: Therefore, the smallest positive perfect cube that can be written as the sum of three consecutive integers is \(\boxed{{\color[HTML]{32cb00}27}}\).

\end{small}
\end{tcolorbox}
\caption{
Criticize and refine a problem in the MATH test set. In the attempt, the model made a hallucination in Step 3, then through stepwise CoT critique, the model detected this error, and accordingly refined the solution starting from Step 3 successfully.}
\label{fig:example_MATH}
\end{figure}

\begin{table}[!hb]
\centering
\begin{tabular}{p{0.95\textwidth}}
\toprule
\textbf{Prompt} \\ 

\midrule
How do you evaluate the following attempt with respect to the problem?
\\
\\
\texttt{<problem>}
\\
\{problem\}
\\
\texttt{</problem>}
\\
\\
\texttt{<attempt>}
\\
\{attempt\}
\\
\texttt{</attempt>}
\\
\\

\texttt{-----} \\
**Notes**: \\
 - Please think step by step. \\
 - Your reasoning should precede any claims or conclusions you make to avoid unwarranted assertions. \\
 - At the end of the evaluation for each step, YOU MUST articulate the conclusion using the format ''Conclusion: Step [i] is correct" or ''Conclusion: Step [i] is incorrect". Words like ''partially correct" are prohibited. \\
 - You shall not evaluate multiple steps at a time, so words like "Step 7 to Step 24:" or "Step 4 through 6" are forbidden. \\
 - Once a mistake is identified and stated, stop the evaluation, and enumerate the corrected steps starting from the step where the mistake was detected, and label this part of your response with \texttt{<correction>} at the start and \texttt{</correction>} at the end. Also, the final answer should be a single number, in the form \textbackslash boxed\{\}, at the final step.
\\ 
\bottomrule
\end{tabular}
\caption{The prompt for the collection of critique and refinement on GSM8K, using GPT4-Turbo.}
\label{table:prompt_gsm8k}
\end{table}

\begin{table}[!hb]
\centering
\begin{tabular}{p{0.95\textwidth}}
\toprule
\textbf{Prompt} \\ 

\midrule
How do you evaluate the following attempt with respect to the problem, with the help of reference solution?
\\
Hint: There could be a mistake.
\\
\\
\texttt{<problem>}
\\
\{problem\}
\\
\texttt{</problem>}
\\
\\
\texttt{<reference\_solution>}
\\
\{reference\_solution\}
\\
\texttt{</reference\_solution>}

\\
\texttt{<attempt>}
\\
\{attempt\}
\\
\texttt{</attempt>}
\\
\\

\texttt{-----} \\
**Notes**: \\
 - Please think step by step. \\
 - Your reasoning should precede any claims or conclusions you make to avoid unwarranted assertions. \\
 - Please ensure that the output text does not include phrases implying the use of a reference solution or hint, even though these resources are being utilized. \\
 - At the end of the evaluation for each step, YOU MUST articulate the conclusion using the format ''Conclusion: Step [i] is correct" or ''Conclusion: Step [i] is incorrect". Words like ''partially correct" are prohibited. \\
 - You shall not evaluate multiple steps at a time, so words like "Step 7 to Step 24:" or "Step 4 through 6" are forbidden. \\
 - Once a mistake is identified and stated, stop the evaluation, and enumerate the corrected steps starting from the step where the mistake was detected, and label this part of your response with \texttt{<correction>} at the start and \texttt{</correction>} at the end. Also, the final answer should be in the form \textbackslash boxed\{\}, at the final step.
\\ 
\bottomrule
\end{tabular}
\caption{The prompt for the collection of critique and refinement on MATH incorrect attempt, using GPT4-Turbo.}
\label{table:prompt_math_incorrect}
\end{table}

\begin{table}[!hb]
\centering
\begin{tabular}{p{0.95\textwidth}}
\toprule
\textbf{Prompt} \\ 

\midrule
For the following text, remove any phrases like "reference solution" or "hint", and keep all the other content. Do not miss the ''\texttt{<correction>}" and  ''\texttt{</correction>}" labels that exist in the text. Do not respond to anything else. \\
\\
\texttt{-----} \\

\{critique\_refinement\}
\\ 
\bottomrule
\end{tabular}
\caption{The prompt for removing the hint of critique and refinement on MATH, using GPT4-Turbo.}
\label{table:prompt_math_incorrect_remove_hint}
\end{table}

\begin{table}[!hb]
\centering
\begin{tabular}{p{0.95\textwidth}}
\toprule
\textbf{Prompt} \\ 

\midrule
How do you evaluate the following attempt with respect to the problem?
\\
Hint: All the steps are correct, and the attempt reached a correct answer.
\\
\\
\texttt{<problem>}
\\
\{problem\}
\\
\texttt{</problem>}
\\
\\
\texttt{<attempt>}
\\
\{attempt\}
\\
\texttt{</attempt>}
\\
\\

\texttt{-----} \\
**Notes**: \\
 - Please think step by step. \\
 - Your reasoning should precede any claims or conclusions you make to avoid unwarranted assertions. \\
 - Please ensure that the output text does not include phrases implying the use of a reference solution or hint, even though these resources are being utilized. \\
 - At the end of the evaluation for each step, YOU MUST articulate the conclusion using the format ''Conclusion: Step [i] is correct" or ''Conclusion: Step [i] is incorrect". Words like ''partially correct" are prohibited. \\
 - You shall not evaluate multiple steps at a time, so words like "Step 7 to Step 24:" or "Step 4 through 6" are forbidden. \\
 - Once a mistake is identified and stated, stop the evaluation, and enumerate the corrected steps starting from the step where the mistake was detected, and label this part of your response with \texttt{<correction>} at the start and \texttt{</correction>} at the end. Also, the final answer should be in the form \textbackslash boxed\{\}, at the final step.
\\ 
\bottomrule
\end{tabular}
\caption{The prompt for the collection of critique on MATH correct attempt, using GPT4-Turbo.}
\label{table:prompt_math_correct}
\end{table}

\begin{table}[!hb]
\centering
\begin{tabular}{p{0.95\textwidth}}
\toprule
\textbf{Prompt} \\ 

\midrule
\#\# Problem \\
\{problem\}
\\
\\
\texttt{-----} \\
Solve the problem step by step, marking each step as "Step [i]:".\\
Your final answer should be in the form \textbackslash boxed\{answer\}, at the end of your response.\\ 
\bottomrule
\end{tabular}
\caption{The prompt for solution sampling.}
\label{table:prompt_solve}
\end{table}

\begin{table}[!hb]
\centering
\begin{tabular}{p{0.95\textwidth}}
\toprule
\textbf{Prompt} \\ 

\midrule
How do you evaluate the following attempt with respect to the problem?
\\
\\
\texttt{<problem>}
\\
\{problem\}
\\
\texttt{</problem>}
\\
\\
\texttt{<attempt>}
\\
\{attempt\}
\\
\texttt{</attempt>}
\\
\\
\texttt{-----} \\
**Notes**: \\
 - Please think step by step. \\
 - Your reasoning should precede any claims or conclusions you make to avoid unwarranted assertions. \\
 - At the end of the evaluation for each step, YOU MUST articulate the conclusion using the format ''Conclusion: Step [i] is correct" or ''Conclusion: Step [i] is incorrect". Words like ''partially correct" are prohibited. \\
\bottomrule
\end{tabular}
\caption{The prompt for critique.}
\label{table:prompt_critic}
\end{table}

\begin{table}[!hb]
\centering
\begin{tabular}{p{0.95\textwidth}}
\toprule
\textbf{Prompt} \\ 

\midrule
How do you refine the following attempt with respect to the problem, given the criticism?\\
\\
\texttt{<problem>}
\\
\{problem\}
\\
\texttt{</problem>}
\\
\\
\texttt{<attempt>}
\\
\{attempt\}
\\
\texttt{</attempt>}
\\
\\
\texttt{<criticism>}
\\
\{wrong\_step\_criticism\}
\\
\texttt{</criticism>}
\\
\bottomrule
\end{tabular}
\caption{The prompt for refinement.}
\label{table:prompt_refine}
\end{table}

\newpage

\begin{table}[!hb]
\centering
\begin{tabular}{p{0.95\textwidth}}
\toprule
\textbf{Prompt of Critique under Critic-CoT} \\ 
\midrule
How do you evaluate the following attempt with respect to the problem?
\\
\\
\texttt{<problem>}
\\
Consider the geometric sequence $\frac{125}{9}, \frac{25}{3}, 5, 3, \ldots$. What is the eighth term of the sequence? Express your answer as a common fraction.
\\
\texttt{</problem>}
\\
\\
\texttt{<attempt>}
\\
Step 1: Identify the first term of the sequence.

The first term \(a_1\) is:
\[
a_1 = \frac{125}{9}
\]

Step 2: Calculate the common ratio \(r\).

The common ratio \(r\) can be found by dividing the second term \(a_2\) by the first term \(a_1\):
\[
r = \frac{a_2}{a_1} = \frac{\frac{25}{3}}{\frac{125}{9}} = \frac{25}{3} \times \frac{9}{125}
\]

Calculating the right-hand side:
\[
= \frac{25 \cdot 9}{3 \cdot 125} = \frac{225}{375}
\]
Now simplify \(\frac{225}{375}\):
\[
= \frac{225 \div 75}{375 \div 75} = \frac{3}{5}
\]

Thus, the common ratio is:
\[
r = \frac{3}{5}
\]

Step 3: Use the formula for the \(n\)-th term of a geometric sequence.

The \(n\)-th term of a geometric sequence can be calculated using the formula:
\[
a_n = a_1 \cdot r^{n-1}
\]

Step 4: Calculate the eighth term \(a_8\).

Substituting \(n = 8\) into the formula:
\[
a_8 = a_1 \cdot r^{8-1} = \frac{125}{9} \cdot \left(\frac{3}{5}\right)^7
\]

Step 5: Calculate \(\left(\frac{3}{5}\right)^7\).

Calculating \(\left(\frac{3}{5}\right)^7\):
\[
\left(\frac{3}{5}\right)^7 = \frac{3^7}{5^7} = \frac{2187}{78125}
\]

Step 6: Substitute back into the expression for \(a_8\).

Now we compute:
\[
a_8 = \frac{125}{9} \cdot \frac{2187}{78125}
\]

Step 7: Multiply the fractions.

\\
\bottomrule
\end{tabular}
\end{table}

\begin{table}
\begin{tabular}{p{0.95\textwidth}}
\toprule
\textbf{Prompt of Critique under Critic-CoT} \\ 
\midrule

To multiply:
\[
a_8 = \frac{125 \cdot 2187}{9 \cdot 78125}
\]

Calculating the denominator:
\[
9 \cdot 78125 = 703125
\]

Calculating the numerator:
\[
125 \cdot 2187 = 273375
\]

Thus:
\[
a_8 = \frac{273375}{703125}
\]

Step 8: Simplify the fraction \(\frac{273375}{703125}\).

To simplify:
First, calculate the GCD of the numerator and denominator. We can find that both 273375 and 703125 are divisible by 135. 

Calculating:
\[
\frac{273375 \div 135}{703125 \div 135} = \frac{2025}{5205}
\]

Continuing to simplify, both 2025 and 5205 are divisible by 15:
\[
\frac{2025 \div 15}{5205 \div 15} = \frac{135}{347}
\]

This is now fully simplified because 135 and 347 have no common factors.

Final Answer:

Thus, the eighth term of the sequence is:
\[
\boxed{ {\color[HTML]{FF0000}\frac{135}{347} } }
\]
\\
\texttt{</attempt>}
\\
\\
\texttt{-----} \\
**Notes**: \\
 - Please think step by step. \\
 - Your reasoning should precede any claims or conclusions you make to avoid unwarranted assertions. \\
 - At the end of the evaluation for each step, YOU MUST articulate the conclusion using the format ''Conclusion: Step [i] is correct" or ''Conclusion: Step [i] is incorrect". Words like ''partially correct" are prohibited. \\

\midrule
\textbf{Response of Critique under Critic-CoT}
\\
\midrule
Step 1: The first term \(a_1 = \frac{125}{9}\) is identified correctly.

Conclusion: Step 1 is correct.

Step 2: The common ratio \(r\) is calculated as:
\[
r = \frac{a_2}{a_1} = \frac{\frac{25}{3}}{\frac{125}{9}} = \frac{25}{3} \times \frac{9}{125} = \frac{225}{375} = \frac{3}{5}
\]
This calculation is correct.

Conclusion: Step 2 is correct.

\\
\bottomrule
\end{tabular}
\end{table}

\begin{table}
\begin{tabular}{p{0.95\textwidth}}
\toprule
\textbf{Response of Critique under Critic-CoT}
\\
\midrule

Step 3: The formula for the \(n\)-th term is correctly stated as:
\[
a_n = a_1 \cdot r^{n-1}
\]

Conclusion: Step 3 is correct.

Step 4: The eighth term is computed as:
\[
a_8 = \frac{125}{9} \cdot \left(\frac{3}{5}\right)^7
\]
This is correct.

Conclusion: Step 4 is correct.

Step 5: Calculating \(\left(\frac{3}{5}\right)^7\):
\[
\left(\frac{3}{5}\right)^7 = \frac{2187}{78125}
\]
This calculation is correct.

Conclusion: Step 5 is correct.

Step 6: Substituting back:
\[
a_8 = \frac{125}{9} \cdot \frac{2187}{78125}
\]
This is correct.

Conclusion: Step 6 is correct.

Step 7: Multiplying the fractions:
\[
a_8 = \frac{125 \cdot 2187}{9 \cdot 78125}
\]
The denominator calculation is correct:
\[
9 \cdot 78125 = 703125
\]
However, for the numerator:
\[
125 \cdot 2187 = 273375
\]
This is also correct.

Conclusion: Step 7 is correct.

Step 8: The goal is to simplify the fraction \(\frac{273375}{703125}\). The attempt made to find the GCD and simplify.
To simplify the fraction, we need to find the greatest common divisor (GCD) of 273375 and 703125. 

The standard approach involves either prime factorization or using the Euclidean algorithm.

Prime Factorization: \\
- 273375:  \\
   - Dividing by 5: \(273375 \div 5 = 54675\) \\
   - Dividing by 5 again: \(54675 \div 5 = 10935\) \\
   - Dividing by 5 again: \(10935 \div 5 = 2187\) \\
   - The number 2187 is \(3^7\). \\
   - Therefore, \(273375 = 5^3 \cdot 3^7\). \\
 \\
- 703125: \\
   - Dividing by 5: \(703125 \div 5 = 140625\) \\
   - Dividing by 5 again: \(140625 \div 5 = 28125\) \\
   - Dividing by 5 again: \(28125 \div 5 = 5625\) \\

\bottomrule
\end{tabular}
\end{table}

\begin{table}
\begin{tabular}{p{0.95\textwidth}}
\toprule
\textbf{Response of Critique under Critic-CoT}
\\
\midrule

   - Dividing by 5 again: \(5625 \div 5 = 1125\) \\
   - Dividing by 5 again: \(1125 \div 5 = 225\) \\
   - Dividing by 5 again: \(225 \div 5 = 45\) \\
   - Finally, \(45 = 3^2 \cdot 5^1\). \\
   - Therefore, \(703125 = 5^6 \cdot 3^2\). \\
 \\
The GCD can now be derived from the lowest powers of the common factors: \\
- For 5: \(5^3\) \\
- For 3: \(3^2\) \\

Thus, the GCD is:
\[
5^3 \cdot 3^2
\]

Therefore, 273375 and 703125 are not divisible by 135. 

Conclusion: Step 8 is incorrect.
\\
\midrule
\textbf{Prompt of Refinement under Critic-CoT} \\ 
\midrule
How do you refine the following attempt with respect to the problem, given the criticism?\\
\\
\texttt{<problem>}
\\
Consider the geometric sequence $\frac{125}{9}, \frac{25}{3}, 5, 3, \ldots$. What is the eighth term of the sequence? Express your answer as a common fraction.
\\
\texttt{</problem>}
\\
\\
\texttt{<attempt>}
\\
Step 1: Identify the first term of the sequence.

The first term \(a_1\) is:
\[
a_1 = \frac{125}{9}
\]

Step 2: Calculate the common ratio \(r\).

The common ratio \(r\) can be found by dividing the second term \(a_2\) by the first term \(a_1\):
\[
r = \frac{a_2}{a_1} = \frac{\frac{25}{3}}{\frac{125}{9}} = \frac{25}{3} \times \frac{9}{125}
\]

Calculating the right-hand side:
\[
= \frac{25 \cdot 9}{3 \cdot 125} = \frac{225}{375}
\]
Now simplify \(\frac{225}{375}\):
\[
= \frac{225 \div 75}{375 \div 75} = \frac{3}{5}
\]

Thus, the common ratio is:
\[
r = \frac{3}{5}
\]

Step 3: Use the formula for the \(n\)-th term of a geometric sequence.

The \(n\)-th term of a geometric sequence can be calculated using the formula:
\[
a_n = a_1 \cdot r^{n-1}
\]

\\
\bottomrule
\end{tabular}
\end{table}

\begin{table}
\begin{tabular}{p{0.95\textwidth}}
\toprule
\textbf{Prompt of Refinement under Critic-CoT} \\ 
\midrule

Step 4: Calculate the eighth term \(a_8\).

Substituting \(n = 8\) into the formula:
\[
a_8 = a_1 \cdot r^{8-1} = \frac{125}{9} \cdot \left(\frac{3}{5}\right)^7
\]

Step 5: Calculate \(\left(\frac{3}{5}\right)^7\).

Calculating \(\left(\frac{3}{5}\right)^7\):
\[
\left(\frac{3}{5}\right)^7 = \frac{3^7}{5^7} = \frac{2187}{78125}
\]

Step 6: Substitute back into the expression for \(a_8\).

Now we compute:
\[
a_8 = \frac{125}{9} \cdot \frac{2187}{78125}
\]

Step 7: Multiply the fractions.

To multiply:
\[
a_8 = \frac{125 \cdot 2187}{9 \cdot 78125}
\]

Calculating the denominator:
\[
9 \cdot 78125 = 703125
\]
Calculating the numerator:
\\

\[
125 \cdot 2187 = 273375
\]

Thus:
\[
a_8 = \frac{273375}{703125}
\]

Step 8: Simplify the fraction \(\frac{273375}{703125}\).

To simplify:
First, calculate the GCD of the numerator and denominator. We can find that both 273375 and 703125 are divisible by 135. 

Calculating:
\[
\frac{273375 \div 135}{703125 \div 135} = \frac{2025}{5205}
\]

Continuing to simplify, both 2025 and 5205 are divisible by 15:
\[
\frac{2025 \div 15}{5205 \div 15} = \frac{135}{347}
\]

This is now fully simplified because 135 and 347 have no common factors.

Final Answer:

Thus, the eighth term of the sequence is:
\[
\boxed{\frac{135}{347}}
\]
\\
\texttt{</attempt>}
\\
\bottomrule
\end{tabular}
\end{table}

\begin{table}
\begin{tabular}{p{0.95\textwidth}}
\toprule
\textbf{Prompt of Refinement under Critic-CoT} \\ 
\midrule
\texttt{<criticism>}
\\
Step 8: The goal is to simplify the fraction \(\frac{273375}{703125}\). The attempt made to find the GCD and simplify.
To simplify the fraction, we need to find the greatest common divisor (GCD) of 273375 and 703125. 

The standard approach involves either prime factorization or using the Euclidean algorithm.

Prime Factorization: \\
- 273375:  \\
   - Dividing by 5: \(273375 \div 5 = 54675\) \\
   - Dividing by 5 again: \(54675 \div 5 = 10935\) \\
   - Dividing by 5 again: \(10935 \div 5 = 2187\) \\
   - The number 2187 is \(3^7\). \\
   - Therefore, \(273375 = 5^3 \cdot 3^7\). \\
 \\
- 703125: \\
   - Dividing by 5: \(703125 \div 5 = 140625\) \\
   - Dividing by 5 again: \(140625 \div 5 = 28125\) \\
   - Dividing by 5 again: \(28125 \div 5 = 5625\) \\
   - Dividing by 5 again: \(5625 \div 5 = 1125\) \\
   - Dividing by 5 again: \(1125 \div 5 = 225\) \\
   - Dividing by 5 again: \(225 \div 5 = 45\) \\
   - Finally, \(45 = 3^2 \cdot 5^1\). \\
   - Therefore, \(703125 = 5^6 \cdot 3^2\). \\
 \\
The GCD can now be derived from the lowest powers of the common factors: \\
- For 5: \(5^3\) \\
- For 3: \(3^2\) \\

Thus, the GCD is:
\[
5^3 \cdot 3^2
\]

Therefore, 273375 and 703125 are not divisible by 135. 

Conclusion: Step 8 is incorrect.
\\
\texttt{</criticism>}
\\
\bottomrule
\end{tabular}
\end{table}
\newpage
\begin{table}[t]
\centering
\begin{tabular}{p{0.95\textwidth}}
\toprule
\textbf{Prompt of Refinement under Critic-CoT} \\ 
\midrule
\texttt{<correction>}
\\
Step 8: Simplify \(\frac{273375}{703125}\). First, we divide it by 375.

Calculating:
\[
\frac{273375 \div 375}{703125 \div 375} = \frac{729}{1875}
\]

Step 9: Further simplify \(\frac{729}{1875}\). The numerator 729 is \(3^6\) and the denominator 1875 is \(3 \cdot 625\). The GCD is 3.

Calculating:
\[
\frac{729 \div 3}{1875 \div 3} = \frac{243}{625}
\]

This fraction is fully simplified.

Final Answer:
Thus, the eighth term of the sequence is:
\[
\boxed{ {\color[HTML]{32cb00}\frac{243}{625}}}
\]
\texttt{</correction>}

\\
\bottomrule
\end{tabular}
\caption{Full example of Critic-CoT. The model got the wrong common factor in Step 8, and through CoT critique, identified this mistake, then took the refinement and reached the correct answer.}
\label{table:Full example of Critic-CoT}
\end{table}

\begin{table}[!hb]
\centering
\begin{tabular}{p{0.95\textwidth}}
\toprule
\textbf{Prompt of Critique under Step-wise Label Critic} \\ 

\midrule
How do you evaluate the following attempt with respect to the problem?
\\
\\
\texttt{<problem>}
\\
\texttt{[...]}
\\
\texttt{</problem>}
\\
\\
\texttt{<attempt>}
\\
\texttt{[...]}
\\
\texttt{</attempt>}
\\
\\
\texttt{-----} \\
**Notes**: \\
 - For each step, you only need to output ''Step [i] is correct" if right, or ''Step [i] is incorrect" if wrong. Do not provide anything else in your output. \\
\midrule
\textbf{Response of Critique under Step-wise Label Critic} \\ 
\midrule
Step 1 is correct.  \\
Step 2 is correct.  \\
Step 3 is correct.  \\
Step 4 is correct.  \\
Step 5 is incorrect.  
\\
\bottomrule
\end{tabular}
\end{table}
\newpage
\begin{table}[t]
\centering
\begin{tabular}{p{0.95\textwidth}}
\toprule
\textbf{Prompt of Refinement under Step-wise Label Critic} \\ 
\midrule
How do you refine the following attempt with respect to the problem, given the criticism?\\
\\
\texttt{<problem>}
\\
\texttt{[...]}
\\
\texttt{</problem>}
\\
\\
\texttt{<attempt>}
\\
\texttt{[...]}
\\
\texttt{</attempt>}
\\
\\
\texttt{<criticism>}
\\
Step 5 is incorrect.
\\
\texttt{</criticism>}
\\
\midrule
\textbf{Response of Refinement under Step-wise Label Critic} \\ 
\midrule
\texttt{<correction>} \\
Step 5: Calculate \(\left(\frac{3}{5}\right)^7\).  
\[
\left(\frac{3}{5}\right)^7 = \frac{3^7}{5^7} = \frac{2187}{78125} \text{ (correct)}
\]

Step 6: Substitute back into the expression for \(a_8\).  
\[
a_8 = \frac{125}{9} \cdot \frac{2187}{78125}
\]

Step 7: Multiply the fractions.  
Calculating the numerator:  
\[
125 \cdot 2187 = 273375 \text{ (correct)}  
\]
Calculating the denominator:  
\[
9 \cdot 78125 = 703125 \text{ (correct)}
\]  

Step 8: Simplify the fraction \(\frac{273375}{703125}\).  
To simplify:  
Calculating the GCD shows that both 273375 and 703125 are divisible by 675, so:  
\[
\frac{273375 \div 675}{703125 \div 675} = \frac{405}{1045}  
\]
Continuing to simplify, both 405 and 1045 are divisible by 5:  
\[
\frac{405 \div 5}{1045 \div 5} = \frac{81}{209}  
\]

Thus, the final answer is:  
\[
\boxed{ {\color[HTML]{FF0000}\frac{81}{209} } }  
\]

\texttt{</correction>}
\\
\bottomrule
\end{tabular}
\caption{Full example of Step-wise Label Critic. The problem and the attempt are the same in Table \ref{table:Full example of Critic-CoT}, so we skip them. Without CoT critique, the model incorrectly believed that Step 5 is wrong, then made the refinement starting from Step 5, which is unsuccessful. }
\label{table:Full example of Step-wise Label Critic}
\end{table}

\begin{table}[!hb]
\centering
\begin{tabular}{p{0.95\textwidth}}
\toprule
\textbf{Prompt of Critique under Outcome Label Critic} \\ 

\midrule
How do you evaluate the following attempt with respect to the problem?
\\
\\
\texttt{<problem>}
\\
\texttt{[...]}
\\
\texttt{</problem>}
\\
\\
\texttt{<attempt>}
\\
\texttt{[...]}
\\
\texttt{</attempt>}
\\
\\
\texttt{-----} \\
**Notes**: \\
 - For the whole attempt, you only need to output ''Each step from Step 1 to Step 8 is correct." if right, or ''Some step from Step 1 to Step 8 is incorrect" if wrong. Do not provide anything else in your output. \\
\midrule
\textbf{Response of Critique under Outcome Label Critic} \\ 
\midrule
Some step from Step 1 to Step 8 is incorrect.
\\
\midrule
\textbf{Prompt of Refinement under Outcome Label Critic} \\ 
\midrule
How do you refine the following attempt with respect to the problem, given the criticism?\\
\\
\texttt{<problem>}
\\
\texttt{[...]}
\\
\texttt{</problem>}
\\
\\
\texttt{<attempt>}
\\
\texttt{[...]}
\\
\texttt{</attempt>}
\\
\\
\texttt{<criticism>}
\\
Some step from Step 1 to Step 8 is incorrect.
\\
\texttt{</criticism>}
\\
\midrule
\textbf{Response of Refinement under Outcome Label Critic} \\ 
\midrule
\texttt{<correction>} \\
Step 1: Identify the first term of the sequence.

The first term \(a_1\) is:
\[
a_1 = \frac{125}{9}
\]

Step 2: Calculate the common ratio \(r\).

The common ratio \(r\) can be found by dividing the second term \(a_2\) by the first term \(a_1\):
\[
r = \frac{a_2}{a_1} = \frac{\frac{25}{3}}{\frac{125}{9}} = \frac{25}{3} \times \frac{9}{125} = \frac{225}{375}
\]
Now simplify \(\frac{225}{375}\):
\[
= \frac{225 \div 75}{375 \div 75} = \frac{3}{5}
\]

\\
\bottomrule
\end{tabular}
\end{table}
\newpage
\begin{table}[t]
\centering
\begin{tabular}{p{0.95\textwidth}}
\toprule
\textbf{Response of Refinement under Outcome Label Critic} \\ 
\midrule

Thus, the common ratio is:
\[
r = \frac{3}{5}
\]

Step 3: Use the formula for the \(n\)-th term of a geometric sequence.

The \(n\)-th term of a geometric sequence can be calculated using the formula:
\[
a_n = a_1 \cdot r^{n-1}
\]

Step 4: Calculate the eighth term \(a_8\).

Substituting \(n = 8\) into the formula:
\[
a_8 = a_1 \cdot r^{8-1} = \frac{125}{9} \cdot \left(\frac{3}{5}\right)^7
\]

Step 5: Calculate \(\left(\frac{3}{5}\right)^7\).

Calculating \(\left(\frac{3}{5}\right)^7\):
\[
\left(\frac{3}{5}\right)^7 = \frac{3^7}{5^7} = \frac{2187}{78125}
\]

Step 6: Substitute back into the expression for \(a_8\).

Now we compute:
\[
a_8 = \frac{125}{9} \cdot \frac{2187}{78125}
\]

Step 7: Multiply the fractions.

To multiply:
\[
a_8 = \frac{125 \cdot 2187}{9 \cdot 78125}
\]

Calculating the denominator:
\[
9 \cdot 78125 = 703125
\]

Calculating the numerator:
\[
125 \cdot 2187 = 273375
\]

Thus:
\[
a_8 = \frac{273375}{703125}
\]

Step 8: Simplify the fraction \(\frac{273375}{703125}\).

To simplify:
Finding the GCD, both 273375 and 703125 are divisible by 675:
\[
\frac{273375 \div 675}{703125 \div 675} = \frac{405}{1042}
\]

Now we check if \(\frac{405}{1042}\) can be simplified further. Both numbers do not have common factors.

Final Answer:

Thus, the eighth term of the sequence is:
\[
\boxed{ {\color[HTML]{FF0000}\frac{405}{1042}}}
\]

\texttt{</correction>}
\\
\bottomrule
\end{tabular}
\caption{Full example of Final Label Critic. The problem and the attempt are also the same in Table \ref{table:Full example of Critic-CoT}. With only the outcome label, the model managed to detect the attempt was incorrect, then made the refinement starting from the first step, which was unsuccessful. }
\label{table:Full example of Final Label Critic}
\end{table}

\newpage

\begin{table}[t]
\centering
\begin{tabular}{p{0.95\textwidth}}
\toprule
\textbf{Prompt of Code Generation on HumanEval} \\ 
\midrule
\#\# Python code problem \\
\begin{lstlisting}[language=Python]
def split_words(txt):
    '''
    Given a string of words, return a list of words split on 
    whitespace, if no whitespaces exists in the text you
    should split on commas ',' if no commas exists you should 
    return the number of lower-case letters with odd order in the
    alphabet, ord('a') = 0, ord('b') = 1, ... ord('z') = 25
    Examples
    split_words("Hello world!") -> ["Hello", "world!"]
    split_words("Hello,world!") -> ["Hello", "world!"]
    split_words("abcdef") == 3 
    '''
\end{lstlisting}

----- \\
Before writing the code, think step by step, marking each step as ``Step [i]:'' \\
Your final function split\_words should be in the form \`{}\`{}\`{}python\textbackslash n[code]\textbackslash n\`{}\`{}\`{}, at the end of your response. 
\\

\midrule
\textbf{Response of Code Generation on HumanEval} \\ 
\midrule
Step 1: Create a function named \`{}split\_words\`{} that takes a string \`{}txt\`{} as an input.

Step 2: Check if there are any whitespace characters in the string. If there are, split the string into words using the `split()` function with the default separator (whitespace).

Step 3: If there are no whitespace characters, check if there are any commas in the string. If there are, split the string into words using the `split()` function with the comma as the separator.

Step 4: If there are neither whitespace characters nor commas in the string, iterate over the characters in the string, keeping track of the count of lowercase letters with odd alphabetical order (based on their ASCII value modulo 2).

Step 5: Return the result based on the conditions met in steps 2, 3, or 4.

Here is the Python code for the function: \\
\`{}\`{}\`{}python 
\begin{lstlisting}[language=Python]
def split_words(txt):
    if ' ' in txt:
        return txt.split()
    elif ',' in txt:
        return txt.split(',')
    else:
        count = sum(1 for char in txt if char.islower() and 
        ord(char) % 2 != 0)
        return count
\end{lstlisting}
\`{}\`{}\`{} \\
This function first checks for whitespace characters, then for commas, and finally counts the lowercase letters with odd alphabetical order if neither of the first two conditions is met. \\

\bottomrule
\end{tabular}
\end{table}
\begin{table}[t]
\centering
\begin{tabular}{p{0.95\textwidth}}

\toprule
\textbf{Prompt of Critique on HumanEval} \\ 
\midrule
How do you evaluate the following attempt with respect to the problem?
\\
<problem> 
\begin{lstlisting}[language=Python]
def split_words(txt):
    '''
    Given a string of words, return a list of words split on 
    whitespace, if no whitespaces exists in the text you
    should split on commas ',' if no commas exists you should 
    return the number of lower-case letters with odd order in the
    alphabet, ord('a') = 0, ord('b') = 1, ... ord('z') = 25
    Examples
    split_words("Hello world!") -> ["Hello", "world!"]
    split_words("Hello,world!") -> ["Hello", "world!"]
    split_words("abcdef") == 3 
    '''
\end{lstlisting}
</problem>
\\
\\
<attempt>
\\
Step 1: Create a function named \`{}split\_words\`{} that takes a string \`{}txt\`{} as an input.

Step 2: Check if there are any whitespace characters in the string. If there are, split the string into words using the `split()` function with the default separator (whitespace).

Step 3: If there are no whitespace characters, check if there are any commas in the string. If there are, split the string into words using the `split()` function with the comma as the separator.

Step 4: If there are neither whitespace characters nor commas in the string, iterate over the characters in the string, keeping track of the count of lowercase letters with odd alphabetical order (based on their ASCII value modulo 2).

Step 5: Return the result based on the conditions met in steps 2, 3, or 4.

Here is the Python code for the function: \\
\`{}\`{}\`{}python 
\begin{lstlisting}[language=Python]
def split_words(txt):
Line 1:    if ' ' in txt:
Line 2:        return txt.split()
Line 3:    elif ',' in txt:
Line 4:        return txt.split(',')
Line 5:    else:
Line 6:        count = sum(1 for char in txt if char.islower() and 
               ord(char) % 2 != 0)
Line 7:        return count
\end{lstlisting}
\`{}\`{}\`{} \\
This function first checks for whitespace characters, then for commas, and finally counts the lowercase letters with odd alphabetical order if neither of the first two conditions is met. \\
</attempt>
\\
\\
\bottomrule
\end{tabular}
\end{table}
\newpage
\begin{table}[t]
\centering
\begin{tabular}{p{0.95\textwidth}}
\toprule
\textbf{Response of Critique on HumanEval} \\ 
\midrule
Step 1: Define a function called \`{}split\_words\`{} that takes a string \`{}txt\`{} as its parameter. \\
Conclusion: Step 1 is correct.

Step 2: Verify if there are any whitespace characters in the string. If they exist, divide the string into words using the \`{}split()\`{} function with the default separator (whitespace).  \\
Conclusion: Step 2 is correct.

Step 3: If no whitespace characters are found, check for commas in the string. If commas are present, split the string into words using the \`{}split()\`{} function with the comma as the separator.  \\
Conclusion: Step 3 is correct.

Step 4: Otherwise, loop through the characters in the string, counting the number of lowercase letters that have an odd alphabetical order (based on their ASCII value modulo 2). The mistake in Step 4 lies in the interpretation of "odd alphabetical order." The problem statement refers to the alphabetical order of lowercase letters in the alphabet, not their ASCII values. The correct approach is to count the lowercase letters whose alphabetical order (where 'a' is 0, 'b' is 1, ..., 'z' is 25) is odd.\\
Conclusion: Step 4 is incorrect.
\\
\midrule
\textbf{Prompt of Refinement on HumanEval} \\ 
\midrule
How do you refine the following attempt with respect to the problem, given the criticism? You shall write another complete Python function, in the format \`{}\`{}\`{}python\textbackslash n[code]\textbackslash n\`{}\`{}\`{}.

\texttt{<problem>}
\\
\texttt{[...]}
\\
\texttt{</problem>}
\\
\\
\texttt{<attempt>}
\\
\texttt{[...]}
\\
\texttt{</attempt>}
\\
\\
<criticism>
\\
Step 4: Otherwise, loop through the characters in the string, counting the number of lowercase letters that have an odd alphabetical order (based on their ASCII value modulo 2). The mistake in Step 4 lies in the interpretation of "odd alphabetical order." The problem statement refers to the alphabetical order of lowercase letters in the alphabet, not their ASCII values. The correct approach is to count the lowercase letters whose alphabetical order (where 'a' is 0, 'b' is 1, ..., 'z' is 25) is odd.\\
Conclusion: Step 4 is incorrect.\\
</criticism>
\\
\\
\bottomrule
\end{tabular}
\end{table}
\newpage
\begin{table}[t]
\centering
\begin{tabular}{p{0.95\textwidth}}
\toprule
\textbf{Response of Refinement on HumanEval} \\ 
\midrule
<correction>
\\

Here is the corrected Python function:

\`{}\`{}\`{}python 
\begin{lstlisting}[language=Python]
def split_words(txt):
    if ' ' in txt:
        return txt.split()
    elif ',' in txt:
        return txt.split(',')
    else:
        count = sum(1 for char in txt if char.islower() 
        and (ord(char) - ord('a')) % 2 != 0)
        return count
\end{lstlisting}
\`{}\`{}\`{} \\

This function correctly counts the lowercase letters with odd alphabetical order by subtracting the ASCII value of 'a' from the ASCII value of each lowercase letter to get its alphabetical order, and then checking if this order is odd. If the order is odd, it increments the count.
\\
</correction>
\\
\bottomrule

\end{tabular}
\caption{Full example of a single-turn refinement on HumanEval. The attempt made mistakes in Step 4 and Line 6, which use the ASCII value directly rather than subtract the value of 'a'. The critique detect this error, and made a successul refinement. }
\label{table:Full example of Critic-CoT on HumanEval}
\end{table}

\end{document}